\documentclass[letterpaper,12pt]{article}

\usepackage{mathptmx}
\usepackage{geometry}

\usepackage{graphicx}

\usepackage[authoryear,comma,longnamesfirst,sectionbib]{natbib}

\usepackage{amsmath, amssymb, amsfonts}
\usepackage{subeqnarray}
\usepackage{algorithm}
\usepackage{algpseudocode}
\usepackage{xcolor}

\usepackage{xr}
\makeatletter
\newcommand*{\addFileDependency}[1]{% argument=file name and extension
  \typeout{(#1)}
  \@addtofilelist{#1}
  \IfFileExists{#1}{}{\typeout{No file #1.}}
}
\makeatother

\newcommand{\prob}[1]{\mathbb{P}\left(#1 \right)}
\newcommand{\hatprob}[1]{\widehat{\mathbb{P}}\left(#1 \right)}

\newcommand{\esp}[1]{\mathbb{E}\left[#1 \right]}
\newcommand{\espi}[2]{\mathbb{E}_{#1}\left[ #2 \right]}
\newcommand{\ind}[1]{I_{\{#1 \}}}
\newcommand{\gp}[1]{\left(#1\right)}
\newcommand{\gacc}[1]{\left\{#1\right\}}

\newtheorem{proposition}{Proposition}

\newcommand{\X}{\mathcal{X}}
\newcommand{\Prob}[1]{\Pr\left( #1 \right)}

\newcommand{\argmax}{\mbox{argmax}}
\newcommand{\gb}[1]{#1}

\newcommand{\blue}[1]{#1}

\newtheorem{definition}{Definition}[section]

\begin{document}

\title{Error rate control for classification rules in multiclass mixture models}

\date{}

\author
{Tristan Mary-Huard\\ 
  GQE - Le Moulon and UMR MIA-Paris\\
  INRAE, CNRS, AgroParisTech, Universit\'e
  Paris-Saclay\\
  %91190, Gif-sur-Yvette, France\\
%, AgroParisTech, INRAE, Universit\'e Paris-Saclay, 75005, Paris, France
\and
Vittorio Perduca\\
Laboratoire MAP5 (UMR CNRS 8145),\\
Universit\'e Paris Descartes, Universit\'e de Paris\\
%75006, Paris, France
\and
Marie-Laure Martin-Magniette\\
Institute of Plant Sciences Paris-Saclay (IPS2) and UMR MIA-Paris\\ %Plant Sciences Paris-Saclay (IPS2) \\
Universit\'e Paris-Saclay, Universit\'e de Paris,\\
AgroParisTech, CNRS, INRAE, Univ. Evry,\\
%91405, Orsay, France. \\
%Universit\'e de Paris, CNRS, INRAE, Institute of Plant Sciences Paris Saclay (IPS2) \\
%91405 Orsay, France.\\
% \\
%Universit\'e Paris-Saclay, CNRS, INRAE, Univ Evry,\\
%91405, Orsay, France. \\
%Universit\'e de Paris, CNRS, INRAE, Institute of Plant Sciences Paris Saclay (IPS2) \\
%%91405 Orsay, France.\\
%UMR MIA-Paris, AgroParisTech, INRAE, Université Paris-Saclay%, 75005, Paris, France. 
\and
Gilles Blanchard\\
Laboratoire de Math\'ematiques d'Orsay,\\
Inria, CNRS, Universit\'e Paris-Saclay
%91405 Orsay, France.
}

\label{firstpage}
\maketitle

\pagebreak

\begin{abstract}
In the context of finite mixture models one considers the problem of classifying as many observations as possible in the classes of interest while controlling the classification error rate in these same classes. Similar to what is done in the framework of statistical test theory, different type I and type II-like classification error rates can be defined, along with their associated optimal rules, where optimality is defined as minimizing type II error rate while controlling type I error rate at some nominal level. It is first shown that finding an optimal classification rule boils down to searching an optimal region in the observation space where to apply the classical Maximum A Posteriori (MAP) rule. Depending on the misclassification rate to be controlled, the shape of the optimal region is provided, along with a heuristic to compute the optimal classification rule in practice. In particular, a multiclass FDR-like optimal rule is defined and compared to the thresholded MAP rules that is used in most applications. It is shown on both simulated and real datasets that the FDR-like optimal rule may be significantly less conservative than the thresholded MAP rule.\\
\end{abstract}

%\begin{keywords}
%Mixture models, classification rule.
%\end{keywords}

\section{Introduction \label{Sect : intro}}

Consider a sample $X_1,\ldots,X_n$ of independent observations in space $\mathcal{X}$ stemming from $P$ populations. The distribution of such a sample can be modelled as a finite mixture of distributions
\begin{eqnarray*}
f(x) = \sum_{p=1}^{P} \pi_p f_p(x)
\end{eqnarray*}
where each class is described by its own probability density distribution $f_p$ and a weight $\pi_p$ so that $0\leq\pi_p\leq1$ and $\sum_p\pi_p=1$. For each observation a label variable $Z$ can be introduced, that equals $p$ if the observation belongs to class $p$. In the following, the labels $Z_1,\ldots,Z_n$ are supposed to be independent, and the mixture model is assumed to be completely known, i.e. the number of classes $P$, the class distributions $f_1,\ldots,f_P$ and the weights $\pi_1,\ldots,\pi_P$ are known.\\

In most practical situations the labels are unobserved, and the goal is to find a suitable clustering of the sample. This is typically obtained by applying a classification rule, i.e. a function $\psi$ that maps $\mathcal{X}$ into $\{1,\ldots,P\}$. The key quantity for building such a function is the posterior probability for observation $x$ to belong to class $p$:
\begin{eqnarray}\label{eq:tau}
\tau_p(x) = \Prob{Z=p|X=x} = \frac{\pi_p f_p(x)}{\sum_{p'=1}^{P} \pi_{p'} f_{p'}(x)}.
\end{eqnarray}
The Maximum A Posteriori (MAP) classification rule, defined as
\begin{eqnarray*}
\psi^{MAP}(x) =  \underset{1\leq p\leq P}{\argmax} \ \tau_p(x)
\end{eqnarray*}
is by far the most popular classification rule. In particular, it is known to minimize the classification error rate $P(\psi(X)\neq Z)$, \cite{McLachPeel00}. \\

While optimality in terms of classification error rate is a desirable property, two drawbacks of the MAP rule should be mentioned. First, optimality does not prevent against a high level of misclassification. When the classification task is difficult, as much as half of the observations can be misclassified by the MAP rule. Second, the MAP rule does not account for the asymmetry that may exist between classes: in some real applications only a small number among the $P$ classes may be of major importance for the experimenter. One should then focus on the misclassification rate on these classes of interest. This situation arises in most cases involving comparisons between different conditions, where one is interested in identifying units having different behaviors in different conditions, but not units whose behavior is unchanged across conditions. For instance in \cite{Berard2011}, a methylation analysis was conducted to identify regions of the genome that are differentially methylated between different organs of Arabidopsis thaliana. This problem can be cast into a clustering problem where units are probes (locations) spread over the Arabidopsis genome, and classes are defined according to the methylation status of these probes in the different organs. Only classes corresponding to differentially methylated behaviors are then of interest.

Different solutions have been proposed to circumvent these two problems. 
One possible approach is to take into account the asymmetry between classes by selecting unequal misclassification cost functions that emphasize the cost of misclassification $c_{k\ell}$ from a non-interesting class $\ell$ into a class $k$ of interest. This has been investigated in \cite{friedman2009elements} for instance. 

%Although the use of partial rules or rules with different misclassification costs is attractive, there exist very few guidelines about how to tune parameters such as $\alpha$ or $c_{k\ell}$ and what will be the impact of this tuning on the final misclassification rate.\\

On the other hand, a high level of misclassification for the MAP rule may come from the fact that it classifies {\em any} observation $x$, regardless of the uncertainty of the classification of point $x$. This uncertainty may be quantified through the probability to not belong to the class $x$ was classified into. A possible solution is to perform a partial classification, using e.g. the thresholded classification rule
\begin{eqnarray}
\psi^{\alpha}(x) = \left\{
\begin{array}{ll}
p &  \text{ if } p =  \underset{1\leq p'\leq P}{\argmax} \ \tau_{p'}(x) \text{ and } \tau_p(x)>1-\alpha,\\
0 &  \text{(not classified) otherwise},
\end{array}
\right. \label{Def : Thresholded rule}
\end{eqnarray}
where $0<\alpha<1$ is a parameter to be chosen by the experimenter. The condition $\tau_p(x)>1-\alpha$ can be reformulated has $1-\tau_p(x)<\alpha$ which amounts to focusing on the classification of observations for which uncertainty is not too high.

\gb{The design of partial rules has been considered through the use of classification rules with a reject (or abstention) option.
An often considered setting there is that the reject option has a fixed cost (less costly than a misclassification error); see
\citet{Chow1970,Herbei06,Pillai13}. This setting has been
also considered in conjunction with specific machine learning
methods such as large margin classifiers \citep{bartlett2008,grandvalet2009,wegkamp2011}, active learning
\citep{zhang2014}, and fair learning \citep{schreuder2021}.}

Alternative partial classification rules include the tight clustering algorithm introduced in \cite{Tseng2005} and extended to deal with large datasets in \cite{Karmakar2019}. In the latter work, when applying the extended tight clustering algorithm to a dataset consisting of more than 50,000 gene expression probes for individuals suffering from psoriasis, more than 30,000 probes were not classified in any of the six clusters that were identified.

%Although the use of partial rules or rules with different misclassification costs is attractive, there exist very few guidelines about how to tune parameters such as $\alpha$ or $c_{k\ell}$, or the cost of abstaining, and what will be the impact of this tuning on the final misclassification rate. 

%An alternative solution for classifying a subset of points in stable and relevant classes is given by the tight clustering algorithm that was introduced in \cite{Tseng2005} and extended to deal with large datasets in \cite{Karmakar2019}. In the latter work, when applying the extended tight clustering algorithm to a dataset consisting of more than 50,000 gene expression probes for individuals suffering from psoriasis, more than 30,000 probes were not classified in any of the six clusters that were identified.

%
%On the other hand, one can take into account the asymmetry between classes by selecting unequal misclassification cost functions that emphasize the cost of misclassification $c_{k\ell}$ from a non-interesting class $\ell$ into a class $k$ of interest. This has been investigated in \cite{friedman2009elements} for instance. 

\gb{All aforementioned approaches (partial rules, rules with different misclassification costs and rule with abstention costs) are attractive, however there exist very few guidelines about how to tune parameters such as $\alpha$, $c_{k\ell}$,  or the cost of abstaining, and what will be the impact of this tuning on the final misclassification rate.}
In this article, we propose a method that addresses both the problem of partial classification and the presence of classes of no interest. 
\gb{In particular, we propose to design the abstention 
region based on a directly interpretable constraint taking the form of an overall statistical confidence level on the effective classifications.}
Based on the control of type I error rate in statistical test theory, we introduce optimal classification rules that guarantee that as many observations as possible are classified, under the constraint that the rate of misclassification in the classes of interest is controlled. As such, our method can be understood as an extension of the Bayesian False Discovery Rate (BFDR) developed in \cite{EfronTibsh02} to general finite mixture models. It is shown that different classification rules should be considered to achieve optimality according to the number $K$ of classes of interest ($K=P$ or $K<P$). Sub-optimality of the thresholded classification rule \eqref{Def : Thresholded rule} is also demonstrated.

\gb{The principle of controlling a specific error rate related to classical statistical confidence criteria has also been considered in, e.g. \cite{Scott05} and \cite{Scott09} in the context of binary classification; %and without abstention option; 
see also \cite{tong2016} for a more recent survey. In that setting, an asymmetry is introduced between the classes: the goal is to maximize correct classification rate in class 1 subject to a fixed control at a prescribed level of the classification error in class 0, either in absolute-value (Neyman-Pearson classification) or in the sense of FDR. 
The setting we introduce here is multi-class and includes 
the possibility of abstention. Furthermore all the classes of interest have the same status (the classes of no interest have a different status, however even if all classes are of interest, our setting makes sense and abstention remains a possibility).
The goal we aim at is to guarantee a certain confidence level over classified examples while minimizing the probability of abstention of classes of interest. The statistical criteria we introduce generalize
the confidence/coverage tradeoff considered in \cite{elyaniv2010,wiener2015,denis2020}, who only considered the binary
classification case where all classes are of interest.}

\gb{The problem of classification with confidence has also been considered under a related but
different angle by~\cite{Lei2014}. In Lei's approach, only the binary classification setting is considered and two regions $C_0$ and $C_1$, where the respective classes are predicted, are constructed such that $C_i$ has a
prescribed coverage $1-\alpha_i$ for class $i$ and the "ambiguous classification region" $C_0\cap C_1$ has minimal overall probability. We observe that the latter overlap region could be interpreted as similar to an abstention region in our setting. However, the criterion used
by~\cite{Lei2014} is different from ours, and finding the optimal classification regions amounts to determine appropriate thresholds for two different level sets of the posterior class probability function. Furthermore, it is unclear how to extend the considered criterion to more than two classes. In our approach, we consider a global constraint which in
our view is more readily interpretable in terms of classification confidence, is naturally adapted to more than two classes, and only necessitates the determination of a single level set of an appropriate function. }
%These former contributions could be connected to the present work to derive optimal classification rules (in the Neyman Pearson or FDR sense) in the context of multiclass supervised classification.}

The paper is organized as follows.  In Section \ref{Sect : Definitions}, notions corresponding to type I and type II error rates are defined in the multiclass classification framework along with optimality for a classification rule. Optimal rules are then exhibited in Section \ref{Sect : Shape}, and heuristic procedures to estimate these rules are presented. The empirical behavior of the procedures is briefly investigated and an application of the proposed strategy to the analysis of differential methylation is presented in Section \ref{Sect : Applications}. Lastly, some discussion is developed in Section \ref{Sect : Discussion}. 

\section{Definitions \label{Sect : Definitions}}

\subsection{Restricted classification rules}
Because the classification task may be arbitrarily difficult for some points of $\mathcal{X}$, we consider restricted rules $\psi_{R}(x)$, that classify points in a subset $R\subseteq\mathcal{X}$ only:
\begin{eqnarray*}
\psi_{R}:\X &\rightarrow& \{0,1,\ldots,P\},
\end{eqnarray*}
where 0 is the status ``not classified'' given to observations in $\overline{R}$, the complementary set of $R$ in $\X$:
$$\forall x\in \bar{R}, \ \psi_{R}(x)=0.$$
As an illustration, consider the thresholded classification rule $\psi^\alpha$ provided in equation \eqref{Def : Thresholded rule}, and define
$$\tau^*(x) = \underset{1\leq p\leq P}{\max}\ \tau_p(x),$$
which corresponds to the maximal posterior probability for point $x$. Then classification rule $\psi^\alpha$ corresponds to the restricted classification rule $\psi_{R_\alpha}$ where
$$R_\alpha=\{x |\tau^*(x)\geq1-\alpha\}.$$

In some cases, the number $K$ of classes of interest may be lower than the total number of classes $P$ in the mixture. In the following, and without loss of generality, we will assume that the classes of interest are the first $K$ classes of the mixture. In this context where $K<P$, it will be convenient to consider classification rules that are also restricted to predict labels in $1,\ldots,K$:
\begin{eqnarray*}
\psi_{K,R}:\X &\rightarrow& \{0,1,\ldots,K\}.
\end{eqnarray*}
In this situation, the maximal posterior probability of interest is
$$
\tau^*_K(x) = \underset{1\leq k\leq K}{\max}\ \tau_k(x).
$$
Note that in the case $K=P$ one has $\tau^*_P(x) = \tau^*(x)$. One can extend the definition of the thresholded classification rule to the case $K<P$ by defining
\begin{eqnarray}
\psi^{\alpha}_K(x) = \left\{
\begin{array}{ll}
k &  \text{ if } k =  \underset{1\leq k'\leq K}{\argmax} \ \tau_{k'}(x) \text{ and } \tau_K^*(x)>1-\alpha,\\
0 &  \text{(not classified) otherwise},
\end{array}
\right. . \label{Def : Restricted Thresholded rule}
\end{eqnarray}

\subsection{Error rates}
A ``good'' classification rule should classify as many observations as possible while controlling the classification error rate. As in the statistical testing setting, one can introduce the type I and type II error rates associated to any given classification rule. We consider two definitions of the type I error rate and one for the type II error rate in the multiclass classification context:
\begin{definition}
Let $\{1,\ldots,K\}$ be the classes of interest (possibly with $K=P$) and $\psi_{K,R}$ be a restricted classification rule. Quantities
\begin{eqnarray*}
MNPR(\psi_{K,R}) &=& \prob{\psi_{K,R}(X)\neq Z, \  \psi_{K,R}(X)>0} = \prob{\psi_{K,R}(X)\neq Z, \  X\in R}\\
\text{and } MFDR(\psi_{K,R}) &=& 
\blue{
\begin{cases}
\prob{\psi_{K,R}(X)\neq Z | \ \psi_{K,R}(X)>0} = \prob{\psi_{K,R}(X)\neq Z | \ X\in R},\\
\qquad \text{ if } \prob{X \in R} >0\\
0, \text{ if } \prob{X \in R}=0,
\end{cases}}
\end{eqnarray*}
are called the multiclass Neyman-Pearson error rate and the multiclass false discovery rate, respectively. The quantity
\begin{eqnarray*}
MFNR(\psi_{K,R}) &=& \prob{ Z \in \{1,\ldots,K\} , \  \psi_{K,R}(X)=0} = \prob{ Z \in \{1,\ldots,K\} , \  X\in \bar{R}}
\end{eqnarray*}
is called the multiclass false negative rate.
\end{definition}
The MFNR quantifies the proportion of observations that belong to a class of interest and that were not classified by rule $\psi_{K,R}$. This quantity corresponds to the false negative proportion in the testing setting, and should be as small as possible. The MNPR and MFDR correspond to possible classification error rates one may want to control at a given level $\alpha$. \gb{Note that a criterion
equivalent to the MFDR has been considered by \citet{elyaniv2010,wiener2015,denis2020}, albeit only in the 
situation $K=P=2$}.

To exemplify the previous definitions, here again \blue{one can consider the thresholded classification rule $\psi^{\alpha}$ ($=\psi^{\alpha}_P$) defined above in the by-default context where all classes are of interest}. One can show that the MFDR of $\psi^{\alpha}$  is always lower than $\alpha$, whatever the number of classes of interest. Indeed one has
\blue{
\begin{eqnarray*}
MFDR(\psi^\alpha) &=& \prob{\psi^\alpha(X)\neq Z | \ X\in R_\alpha}\\
&=& \frac{\prob{\psi^\alpha(X)\neq Z \cap \ X\in R_\alpha}}{\prob{X\in R_\alpha}}.\\
\end{eqnarray*}
The numerator in the last expression can be reformulated as follows:
\begin{eqnarray*}
\prob{\psi^\alpha(X)\neq Z \cap \ X\in R_\alpha} &=& \int_{\mathcal{X}}\espi{Z|X=x}{\ind{\psi^\alpha(x)\neq Z}}\ind{x\in R_\alpha}f(x)dx\\
&=& \int_{\X}\espi{Z|X=x}{Z\neq p^*}\ind{x\in R_\alpha}f(x)dx\\
&=& \int_{\X}(1-\tau^*_P(x))\ind{x\in R_\alpha}f(x)dx
\end{eqnarray*}
where $p^*=\arg\underset{p}{\max}\tau_p(x)$. Plugging this expression into the previous equation leads to
\begin{eqnarray*}
MFDR(\psi^\alpha) &=& \frac{1}{\prob{X\in R_\alpha}}\int_{R_\alpha} (1-\tau^*_P(x))f(x)dx\\
&\leq& \frac{1}{\prob{X\in R_\alpha}}\int_{R_\alpha} \alpha f(x)dx\\
&\leq& \alpha.
\end{eqnarray*}
}

%In the following the upperscript $K$ in $\psi^K_R$ will be skipped for simplicity, and $\psi_R$ should be understood as $\psi^K_R$, with $K=P$ or $K<P$ depending on the context.

%\subsection{Estimation}\label{Subsect : Estimation}
%Assuming one disposes of a $n$-sample $D_n=(X_1,\ldots,X_n)$ (with possibly associated labels $(Z_1,\ldots,Z_n)$ also available), the different error rates defined in the previous section can be estimated for any restricted classification rule by
%\begin{eqnarray*}
%\widehat{MNPR}=\hatprob{ \psi_R(X) \neq Z ,\ \ X\in R } &=& \frac{1}{n}\sum_{X_i \in R} (1-\tau^*(X_i))\\
%\widehat{MFDR}=\hatprob{ \psi_R(X) \neq Z | \ X\in R } &=& \frac{1}{n_R}\sum_{X_i \in R} (1-\tau^*(X_i))\\
%\widehat{MFNR}=\hatprob{ \psi_R(X) = 0  ,\ \ Z\in \{1, \ldots, K\} } &=& \frac{1}{n}\sum_{X_i \in \overline{R}} \sum_{j=1}^K \tau_j(X_i)
%\end{eqnarray*}
%respectively, where $n_R$ is the number of observations in $D_n$ belonging to region $R$. %Importantly, these estimators can be computed for any restricted classification rule, leading to a first way to compare rules between them.

\subsection{Optimal classification rules}
We now introduce a formal definition for the $optimality$ of a restricted classification rule.
\begin{definition}
Given a level $\alpha$ ($0\leq \alpha \leq 1$) and a set of classes of interest  $\{1,\ldots,K\}$ (possibly with $K=P$), a classification rule $\psi_{K,R}^*$ is MNPR-optimal at level $\alpha$ if
\begin{eqnarray}
\psi_{K,R}^* = \arg\underset{\psi_{K,R}}{\min}\  MFNR(\psi_{K,R})\ \  u.c. \ \ MNPR(\psi_{K,R}) \leq \alpha . \label{Equ : NP-optimality}
\end{eqnarray}
Alternatively, $\psi_{K,R}^*$ is MFDR-optimal at level $\alpha$ if
\begin{eqnarray}
\psi_{K,R}^* = \arg\underset{\psi_{K,R}}{\min}\  MFNR(\psi_{K,R})\ \  u.c. \ \ MFDR(\psi_{K,R}) \leq \alpha . \label{Equ : FDR-optimality}
\end{eqnarray}
\end{definition}
Note that the definition implies both a region $R$ as large as possible and an optimal classification at each point $x$ in region $R$.

\section{Optimal classification rules \label{Sect : Shape}}

The goal of the present section is to exhibit the shape of the optimal classification rules for problems \eqref{Equ : NP-optimality} and \eqref{Equ : FDR-optimality}. To this end, we will first prove that for a fixed region $R$ it is optimal to apply the restricted MAP rule. As a consequence, looking for an optimal rule $\psi_{R}^*$ actually boils down to finding the optimal region $R^*$ where to apply the (restricted) MAP rule. The theoretical form of the optimal region will then be derived for problems \eqref{Equ : NP-optimality} and \eqref{Equ : FDR-optimality}. Each optimal region requires the tuning of a unknown threshold $\lambda$, for which we will provide a heuristic estimation method.

\subsection{Optimal rule when $R$ is fixed}
Let us first define the restricted MAP classification rule $\psi^{MAP}_{K,R}$ as
\begin{eqnarray*}
\psi^{MAP}_{K,R}(x) = \left\{
\begin{array}{l}
\arg\underset{1\leq k\leq K}{\max} \tau_k(x) \text{ if } x\in R \\
0 \text{ otherwise.}
\end{array}
\right. 
\end{eqnarray*}

\blue{Let first consider a classification $\psi_{K,R}$ restricted to the region $R$. At any point $x\in R$ one has}
%For any classification rule $\psi_{K,R}$ restricted to the same region $R$ (and the same $K$ classes) and any point $x\in R$ one has
\begin{eqnarray*}
\prob{ \psi_{K,R}(x) \neq Z } &=& \sum_{k=1}^K (1-\tau_k(x))\ind{\psi_{K,R}(x)=k} \\
    &\geq& (1-\tau^*_K(x)) \\
    &\geq& \prob{ \psi^{MAP}_{K,R}(x) \neq Z }
\end{eqnarray*}
and therefore
\begin{eqnarray*}
MNPR(\psi_{K,R}) &\geq& MNPR(\psi^{MAP}_{K,R})\\
\text{ and } MFDR(\psi_{K,R}) &\geq& MFDR(\psi^{MAP}_{K,R}).
\end{eqnarray*}
Alternatively it is straightforward to observe that
\begin{eqnarray*}
MFNR(\psi_{K,R}) = MFNR(\psi^{MAP}_{K,R}).
\end{eqnarray*}
\blue{As a consequence, one can conclude that for a fixed region $R$ the optimal restricted classification rule is $\psi^{MAP}_{K,R}$. Therefore finding the solutions of problems \eqref{Equ : NP-optimality} and \eqref{Equ : FDR-optimality} boils down to finding the region $R^*$ to which the MAP classification rule should be applied.}
%and therefore  each solution of problems \eqref{Equ : NP-optimality} and \eqref{Equ : FDR-optimality} is obtained by applying the MAP classification rule to an optimal classification region $R^*$ to be defined, that is $\psi^*_{K,R}=\psi^{MAP}_{K,R^*}$.

In the following the upperscript $MAP$ and the lowerscript $K$ in $\psi^{MAP}_{K,R}$ will be skipped for simplicity, and $\psi_R$ should be understood as $\psi^{MAP}_{K,R}$, with $K=P$ or $K<P$ depending on the context.

\subsection{Optimal classification region for MNPR control} \label{Subsect : MNPR control}
If all classes are of interest ($K=P$), one looks for a region $R^*$ such
that
\begin{eqnarray*}
R^* \in \underset{R}{\arg\min}\ P\gp{X\in \bar{R}} \ \text{ u.c. } \ \prob{ \psi_R(X) \neq Z , \  X\in R } \leq \alpha .
\end{eqnarray*}
If $K<P$, the region $R^*$ should satisfy
\begin{eqnarray}
\label{eq:MNPRpb}
R^* \in \underset{R}{\arg\min}\ P\gp{Z\in \{1,\ldots,K\}, \  X\in \bar{R}} \ \text{
u.c. } \ \prob{ \psi_R(X) \neq Z , \  X\in R } \leq \alpha .
\end{eqnarray}
%In the following we will assume that the marginal distribution of $X$ is without atoms. 
%Consequently, one can always enforce the equality in the constraints above.
%\gb{In both cases, we can enforce the equality in the constraint, because  the marginal distribution of $X$ is assumed without atoms.}
\begin{proposition}\label{Prop:MNPR classification region}
    \blue{If $MNPR(\psi_{\X})\leq \alpha$, then $R^*=\X$ is a solution
    of ~\eqref{eq:MNPRpb}. Otherwise,
    any $R^*$ satisfying $MNPR(\psi_{R^*}) = \alpha$ and
  % is of the form
  such that
\begin{eqnarray}
\gacc{ \ x \ \left|\
  \frac{\sum_{k=1}^K \tau_k(x)}{1-\tau^*_K(x)}> \lambda \right. } \ \subseteq \ R^* \ \subseteq \ \gacc{ \ x \ \left|\
  \frac{\sum_{k=1}^K \tau_k(x)}{1-\tau^*_K(x)}\geq \lambda \right. }  ,
%  \ \
%\bar{R^*} \subseteq \gacc{ \ x \ \left|\
%  \frac{\sum_{k=1}^K \tau_k(x)}{1-\tau^*_K(x)}\leq \lambda \right. } \ \ ,
  \label{Equ : OptimalRule-NP-KinfP }
\end{eqnarray}
for some $\lambda\geq 0$, is a solution of~\eqref{eq:MNPRpb}.}

\blue{If the marginal distribution of $X$ is without atoms,
there exists $R^*$ satisfying the above conditions.}

In the particular case where $K=P$, %$R^*$ satisfies for some 
\blue{condition~\eqref{Equ : OptimalRule-NP-KinfP } can be reduced to:}
%is of the form
\begin{eqnarray}
\gacc{ \ x \ \left| \
  \tau^*(x) > \lambda \right. } \  \subseteq \
  R^* \ \subseteq \ \gacc{ \ x \ \left| \
  \tau^*(x) \geq \lambda \right. } .
  %; \ \
%\bar{R^*} \subseteq \gacc{ \ x \ \left| \
%  \tau^*(x) \leq \lambda \right. } \ .
  \label{Equ : OptimalRule-NP-KequP }
\end{eqnarray}
\blue{for some $\lambda\geq 0$.}
\end{proposition}
The proof is based on the original proof of the Neyman-Pearson theorem, \cite{neyman1933ix}, and can be found in Appendix A.

\subsection{Optimal classification region for MFDR control} \label{Subsect : MFDR control}
If all classes are of interest, one looks for a region $R^*$ such
that
\begin{eqnarray*}
R^* \in \underset{R}{\arg\min}\ P\gp{X\in \bar{R}} \ \text{ u.c. } \ \prob{ \psi_R(X) \neq Z | \ X\in R } \leq \alpha.
\end{eqnarray*}
If $K<P$, $R^*$ should satisfy
\begin{eqnarray}
R^* \in \underset{R}{\arg\min}\ P\gp{Z\in\{1,\ldots,K\} , \  X\in \bar{R}} \ \text{
u.c. } \ \prob{ \psi_R(X) \neq Z| \ X\in R } \leq \alpha.
\label{eq:MFDRpb}
\end{eqnarray}
\begin{proposition}\label{Prop:MFDR classification region}
\blue{If $MFDR(\psi_{\X}) \leq \alpha$, then $R^*=\X$
  is a solution of~\eqref{eq:MFDRpb}. If $\tau^*_K(X)<1-\alpha$ almost surely, then $R^*=\emptyset$ is a solution of~\eqref{eq:MFDRpb}.}
  
\blue{Otherwise, any $R^*$ satisfying $MFDR(\psi_{R^*}) = \alpha$ and
  % is of the form
  such that
%$R^*$ satisfies
\begin{eqnarray}
  \gacc{ \ x \ \left|\
  \frac{1-\alpha-\tau^*_K(x)}{\sum_{k=1}^K\tau_k(x)}< \lambda \right. }
  \ \ \subseteq \ \ R^*  \ \ \subseteq \ \
\gacc{ \ x \ \left|\
\frac{1-\alpha-\tau^*_K(x)}{\sum_{k=1}^K\tau_k(x)} \leq \lambda \right. } , \label{Equ : OptimalRule-FDR-KinfP}
\end{eqnarray}
for some $\lambda\geq 0,$ is a solution of~\eqref{eq:MFDRpb};
if the marginal distribution of $X$ is without atoms, such an $R^*$
exists.}

\blue{In the case $K=P$, 
%$R^*$ satisfies 
condition~\eqref{Equ : OptimalRule-FDR-KinfP} can be reduced to:}
\begin{eqnarray}
\gacc{ \ x \ \left| \ \tau^*(x)>
  1-\alpha-\lambda \right. }
  \ \ \subseteq \ \ R^* \ \ \subseteq \ \
  \gacc{ \ x \ \left| \ \tau^*(x) \geq
  1-\alpha-\lambda \right. } \ ,
  \label{Equ : OptimalRule-FDR-KequP}
\end{eqnarray}
\blue{for some $\lambda \geq 0$.}
\end{proposition}
The proof of Proposition \ref{Prop:MFDR classification region} can be found in Appendix B. \gb{In the case $K=P$, 
the optimal region takes the form of a thresholded rule~\eqref{Def : Thresholded rule} for an appropriate threshold. This was
observed by \cite{denis2020} who studied optimal rules for
the MFDR/coverage tradeoff in the case
$K=P=2$. Interestingly, when $K<P$, the optimal rule is 
different from a simple threshold rule.}

The form of the optimal region $R^*$ in~\eqref{Equ : OptimalRule-NP-KequP },~\eqref{Equ : OptimalRule-FDR-KinfP}  is $C_{<\lambda} \subseteq R^* \subseteq C_{\leq \lambda}$, where $C_{<\lambda},C_{\leq\lambda}$ are sublevel sets  (in the sense of strict, resp. non-strict inequality)  of a criterion depending on the posterior probabilities at point $x$. This ``sandwiching'' relation is  theoretically relevant in situations where $\prob{C_{<\lambda}} < \prob{C_{\leq \lambda}}$, for example if  the criterion used is piece-wise constant, and the level set at value $\lambda$ has nonzero probability.  In this situation, strictly speaking to achieve exactly the target MNPR or MFDR rate, $R^*$ should only include  part of the level set at value $\lambda$ (and this part can be chosen arbitrarily provided the constraint is satisfied). \blue{The assumption of atom-free $X$-marginal could also be lifted to grant the existence of optimal rules in general, provided randomized rules are allowed, as in classical Neyman-Peason theory.} Since such situations are obviously of little relevance for most applications, in the remainder of this work, to simplify exposition we assume that $\prob{C_{<\lambda}} = \prob{C_{\leq \lambda}}$ and that $R^*$ exactly coincides with a sublevel set.

\subsection{Comparison between thresholded rule and the MFDR-optimal rule}
The restriction region $R_\alpha$ of the thresholded rule \eqref{Def : Restricted Thresholded rule} can now be compared to the MFDR-optimal regions $R^*$ found in the previous section. One has
\begin{eqnarray*}
R_\alpha &=& \{x \left| \tau^*_K(x)\geq1-\alpha \right.\} \text{ with } K \leq P, \\
%R^* &=& \{x \left| \tau^*(x)\geq \lambda\right.\} \text{ with } \lambda\leq1-\alpha \text{ if } K=P,\\
%R^* &=& \left\{x | \tau^*_K(x)\geq 1-\alpha -\lambda\sum_{j=1}^K\tau_j(x) \right\} \text{ with } \lambda\geq0 \text{ if } K<P.
%\end{eqnarray*}
%R^* & = & \left\{
%\begin{array}{l}
%\left\{x | \tau^*_K(x)\geq 1-\alpha - \lambda\sum_{j=1}^K\tau_j(x) \right\} \text{ with } \lambda \geq 0 \text{, if } K<P\\
%\{x \left| \tau^*(x)\geq 1-\alpha-\lambda\right.\} \text{ with } \lambda \geq 0 \text{, if } K=P
%\end{array}
%\right.
R^* &=& \left\{x | \tau^*_K(x)\geq 1-\alpha - \lambda\sum_{k=1}^K\tau_k(x) \right\} \text{ with } \lambda \geq 0 \text{, if } K \leq P.
\end{eqnarray*}

%The form of the restriction region when $K<P$ can be interpreted as follows: since the threshold for $\tau_K^*(x)$ decreases a soon as $\lambda\sum_{j=1}^K\tau_j(x)$ is high, it is worth to attempt to classify observation $x$ if it is likely that $x$ belongs to one of the populations of interest.\\
According to the shape of $R^*$, the threshold for classifying an observation $x$ based on the maximal posterior probability $\tau^*_K(x)$ depends not only on $\alpha$, but also on $\sum_{k=1}^K\tau_k(x)$, that is the probability of belonging to the overall group of populations of interest. When $K<P$ this can be interpreted as follows:
when $\tau^*_K(x)$ is mild, the optimal rule may still classify $x$ if the probability of $x$ to belong to the overall group of the populations of interest is high. This interesting feature is not accounted for
%if it is likely that an observation $x$ belongs to one of the populations of interest, then the threshold for $\tau_K^*(x)$ is low, and it is therefore worth to classify $x$, even though the posterior probability of each population of interest taken individually is small.
in the thresholded rule that classifies an observation only based on $\tau^*_K(x)$. In this sense, the thresholded rule is not based on the appropriate classification criterion when one is interested only in a subset of all possible populations. \\
More generally, one can observe that whatever the number of populations of interest, both $\psi_{R^*}$ and $\psi^\alpha$ classify observations based on $\tau^*(x)$, but using a threshold that is always lower for $\psi_{R^*}$. \blue{Consequently i) the MFNR will always be higher for $\psi^\alpha$, and ii) the MFDR of $\psi^\alpha$ will be lower than the requested nominal level, i.e. the thresholded rule is a \emph{conservative} classification strategy to control the MFDR at level $\alpha$}. In Section \ref{Sect : Applications} we illustrate on simulated and real data the fact that the gap in term of MFNR between $\psi_{R^*}$ and $\psi^\alpha$ may be high.\\

%\subsection{Estimation of parameter $\lambda$}
%Error rate control as presented in the previous section requires the knowledge of the class distributions $f_1,\ldots,f_P$ and their associated weights $\pi_1,\ldots,\pi_P$, since both the conditional probabilities and threshold $\lambda$ depend on them.
%In real experiments these quantities are partially or completely unknown, and have to be inferred. In some cases (discriminant analysis or parametric mixture models for instance), the class distributions are known up to some parameters, and the distribution inference reduces to parameter inference. One can then use classical inference methods (maximum likelihood or EM algorithm) to access estimates $\widehat{f}_0,\ldots,\widehat{f}_K$ and consequently estimates of conditional probabilities and $\lambda$. Alternatively, in some cases (for instance logistic regression) only the conditional probabilities can be inferred. Since in all cases one obtains estimates of the conditional probabilities, we propose here a heuristic strategy to evaluate threshold $\lambda$ of the optimal region based on these estimated conditional probabilities. Notice that so far we did not distinguish between supervised and unsupervised classification: the definitions and results proposed so far are meaningful in both contexts along with the following heuristic. \\

\subsection{Estimation of parameter $\lambda$}\label{Subsect: EstimationOfLambda}

We present here an heuristic strategy to choose $\lambda$ from the data at hand.
% and obtain the optimal region $R^*$.
First note that all optimal regions defined in Sections~\ref{Subsect : MNPR control} and~\ref{Subsect : MFDR control} are of the form of level sets
$$R^* = \gacc{ \ x \ \left|\
Crit(x) > \lambda \right. }, $$ where  $Crit(x)$ is a criterion based on the posterior probabilities at point $x$, that depends on the risk one wants to control and on the number of classes of interest as follows:\\ %(up to the sign of $\lambda$ for Equation \eqref{Equ : OptimalRule-FDR-KinfP}), as
$\star$ if one aims at controlling the MNPR then
$$ Crit(x) = \frac{\sum_{k=1}^K \tau_k(x)}{1-\tau^*_K(x)}\ (\text{case } K<P )\quad \text { or }\quad Crit(x) =\tau^*(x) \ (\text{case } K=P), $$
$\star$ if one aims at controlling the MFDR then
$$ Crit(x) = \frac{\tau^*_K(x)+\alpha-1}{\sum_{k=1}^K\tau_k(x)}\ (\text{case } K<P )\quad \text { or }\quad Crit(x) =\tau^*(x)+\alpha-1 \ (\text{case } K=P) $$
where $\lambda\leq 0$.

%\begin{table}
%%\label{Tab:Crit}
%\begin{tabular}{lll}
%                 & $K<P$ & $K = P$ \\
%\hline
%$MNPR(\psi_{R^*})$  & $\gacc{ \ x \ \left|\
%\frac{\sum_{k=1}^K \tau_k(x)}{1-\tau^*_K(x)}> \lambda \right. }$ & $\gacc{ \ x \ \left| \ \tau^*(x)> \lambda \right. }$ \\
%\hline
%$MFDR(\psi_{R^*})$  &  $\gacc{ \ x \ \left|\
%\frac{\tau^*_K(x)+\alpha-1}{\sum_{k=1}^K\tau_k(x)}> \lambda \right. },\, \lambda\leq 0$ & $\gacc{ \ x \ \left| \ \tau^*(x)+\alpha-1>
%\lambda \right. },\, \lambda\leq 0$
%\end{tabular}
%\bigskip
%\bigskip
%\caption{Shape of the optimal region $R^*$ when controlling the $MNPR$ or $MFDR$ at level $\alpha$ with $K$ out of $P$ classes of interest.}
%\label{Tab:Crit}
%\end{table}

%One aims at controlling MFDR or MNPR.
Assuming an $n$-sample $D_n=(X_1,\ldots,X_n)$ is available, the MNPR and MFDR can be estimated for any region $R$ by
\begin{eqnarray*}
&&\hatprob{ \psi(X) \neq Z ,\ \ X\in R } = \frac{1}{n}\sum_{X_i \in R} (1-\tau^*_K(X_i))\\
\text{ and }&& \hatprob{ \psi(X) \neq Z | \ X\in R } = \frac{1}{n_R}\sum_{X_i \in R} (1-\tau^*_K(X_i))
\end{eqnarray*}
respectively, where $n_R$ is the number of observations in $D_n$ belonging to $R$ (and we recall that $\tau^*_K(x)=\tau^*(x)$ if $K=P$). This leads to the following general heuristic for the evaluation of threshold $\lambda$:
% Code pour afficher l'algorithme si on utilise le package algorithmic, ce qui ne semble pas marcher ici:
%\begin{algorithm}[H]
%\caption{}
%\label{Algo: Forward}
%\begin{algorithmic}
%\REQUIRE $\alpha$, $x_1,\ldots,x_n$
%\STATE 1/ Order observations $x_1,\ldots,x_n$ according to $Crit(x_i)$:\\ \vspace{0.3cm}
%\hspace{2cm}$Crit(x_{(1)})\geq\ldots\geq Crit(x_{(N)})$,
%\STATE 2/ Find the largest index $i_{max}$ such that $\displaystyle{\frac{1}{M(i_{max})}\sum_{i=1}^{i_{max}} (1-\tau^*(x_{(i)})) \leq \alpha}$,
%\STATE 3/ Set $\widehat{\lambda} = 1-\tau^*(x_{(i_{max})})$,
%\STATE 4/ Output $\widehat{\lambda}$.
%\end{algorithmic}
%\end{algorithm}
% Code en utilisant le package algpseudocode:
\begin{algorithm}[H]
%\caption{}
%\label{Algo: Forward}
\begin{algorithmic}
\Require $\alpha$, $x_1,\ldots,x_n$
\State 1/ Order observations $x_1,\ldots,x_n$ according to $Crit(x_i)$:\\ \vspace{0.3cm}
\hspace{2cm}$Crit(x_{(1)})\geq\ldots\geq Crit(x_{(N)})$,
\State 2/ Find the largest index $i_{max}$ such that $\displaystyle{\frac{1}{M(i_{max})}\sum_{i=1}^{i_{max}} (1-\tau^*_K(x_{(i)})) \leq \alpha}$,
\State 3/ Set $\widehat{\lambda} = Crit(x_{(i_{max})})$, %#$\widehat{\lambda} = 1-\tau^*_K(x_{(i_{max})})$,
\State 4/ Output $\widehat{\lambda}$.
\end{algorithmic}
\caption{}
\label{Algo: Forward}
\end{algorithm}
If the goal is to control MFDR, then in the second step of the heuristic $M(i_{max})=i_{max}$, otherwise $M(i_{max})=n$ for a MNPR control. %The choice of the criterion depends on the quantity to control and the number of classes of interest, and is deduced from the optimal regions given in sections \ref{Subsect : MNPR control} and \ref{Subsect : MFDR control}.

\section{Applications \label{Sect : Applications}}

The aim of this section is to illustrate the performance of the MFDR classification rule derived in the previous section. We first consider different scenarios based on simulated data. In a first scenario the true posterior probabilities are available, which corresponds to the theoretical setting of Sections~\ref{Sect : Definitions} and~\ref{Sect : Shape}. In this scenario the only parameter to be estimated is $\lambda$, and we empirically evaluate the ability of the previous algorithm %\eqref{Algo: Forward}
to efficiently estimate $\lambda$ and control the error rate at the required nominal level. The performance of the optimal rule is also compared to the performance of the thresholded rule. In a second scenario the true posterior probabilities are assumed to be unknown - the by-default setting of most application cases. This scenario allows us to evaluate the impact of estimating the posterior probabilities on the error rate control procedure. Lastly, we present an application on real data, where the MFDR control procedure is applied to the differential analysis of methylation profiles.

\subsection{Simulation setting}
Datasets are simulated from a mixture of 3 bidimensional Gaussian distributions whose mean vectors are $(-1,0)$, $(0,D)$ and $(1,0)$, respectively. Here $D$ is a parameter that tunes the distance between populations 1 and 3 and population 2. When $D=0$ the three populations are highly overlapping, whereas a high value of $D$ makes the populations more distinct. In each class the covariance matrix is diagonal, with an identical variance $\sigma^2$. Weights $\pi_1,\pi_2,\pi_3$ are all fixed at 1/3. For each dataset, 200 observations per population are simulated. In this context the easiness of the classification task is ruled by parameters $D$ and $\sigma^2$: the higher $D$ (respectively the lower $\sigma^2$), the easier the classification. Several values are considered for $D$ (0, 1, 2, 3) and $\sigma^2$ (0.5, 1, 2). An illustration of different configurations ranging from an ``easy'' classification case ($D=3$, $\sigma^2=0.5$) to a ``hard'' one ($D=0$, $\sigma^2=2$) can be found in Appendix C. Finally, 100 datasets are generated for each configuration.

\blue{In a supplementary simulation study, datasets were generated from a mixture of three bivariate Student distributions; details can be found in Appendix F.}

% Commenting out to solve issues with the display of the figure
%\begin{figure}
%\caption{Three examples of simulated data, with different parameter values: an "easy" case (left) corresponding to $D=3$ and $\sigma^2=0.5$, and intermediate case (center) corresponding to $D=2$ and $\sigma^2=1$ and a difficult case (right) corresponding to  $D=0$ and $\sigma^2=2$. Colors correspond to class labels. \label{fig:data}}
%  \centering
% \begin{tabular}{ccc}
% \includegraphics[scale=0.25]{D:/Dropbox/MFDR/Resultats/2019/Easy.pdf}&
% \includegraphics[scale=0.25]{D:/Dropbox/MFDR/Resultats/2019/Intermediate.pdf}
% \includegraphics[scale=0.25]{D:/Dropbox/MFDR/Resultats/2019/Difficult.pdf}
% \end{tabular}
%\end{figure}

\subsection{\blue{Case 1: posterior probabilities are known}}

In what follows the objective is to perform classification with a MFDR controlled at a nominal threshold fixed at $\alpha=0.05$.

In Section 2.2, a straightforward calculation has shown that
the thresholded classification rule $\psi_\alpha$ with a threshold fixed at $1-\alpha$ guarantees
a MFDR control at a level $\alpha$. However the results of Section 3.3 suggest that fixing the threshold at $1-\alpha$ may result in a high MFNR. The alternative procedure for the optimal rule described in Section \ref{Subsect: EstimationOfLambda} should yield better performance thanks to a less conservative choice for the threshold. The potential gain (in terms of MFNR) of the optimal rule compared to the $1-\alpha$ thresholded rule is investigated in this section.

\begin{figure}
  \centering
\includegraphics[scale=0.7]{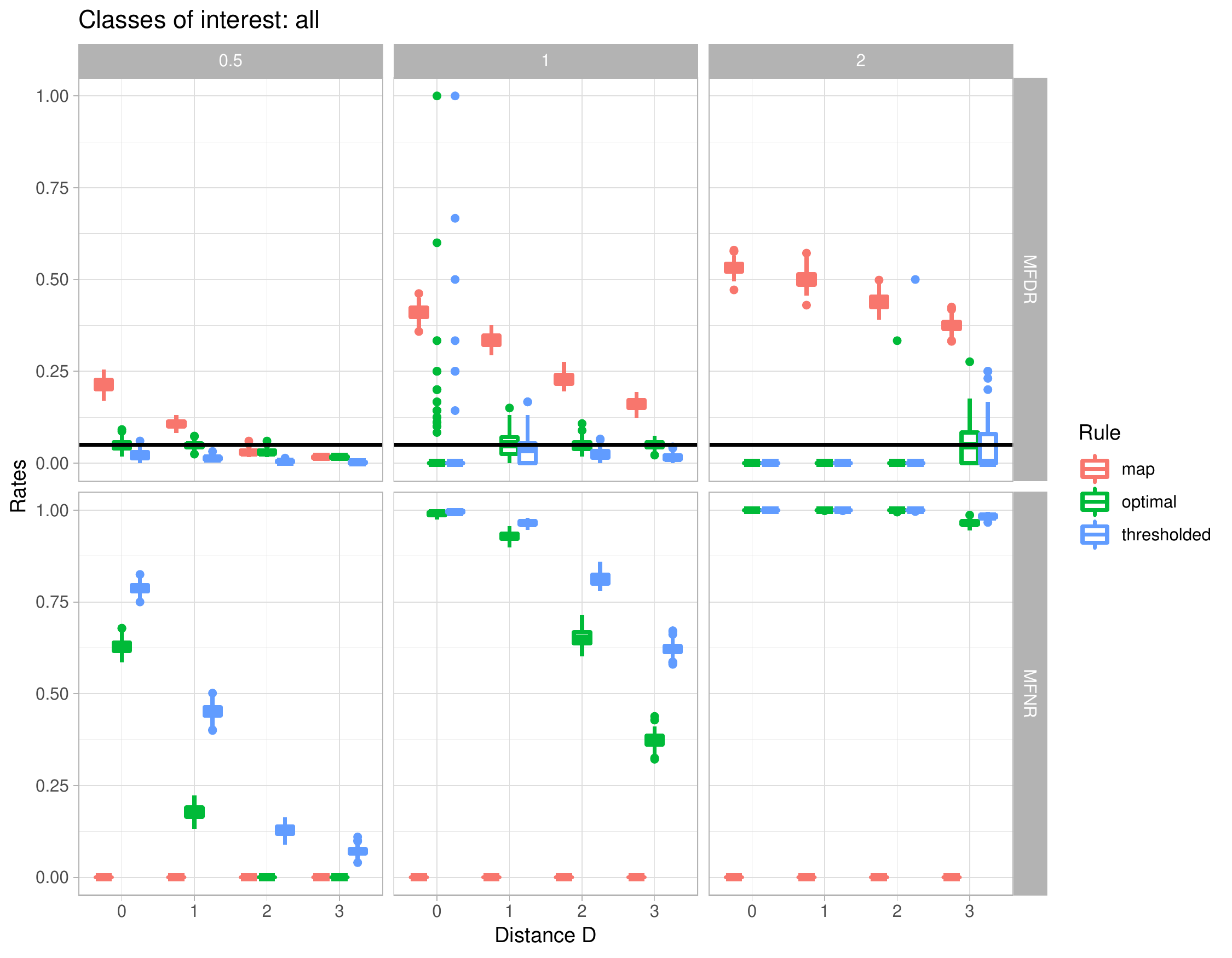}  
\caption{ Performances of the MAP, optimal and $1-\alpha$ thresholded classification rules in terms of realized MFDR and MFNR. Columns correspond to different values of the variance $\sigma^2$. The black line on the top graphs corresponds to the nominal level of 5\%. This figure appears in color in the electronic version of this article, and color refers to that version.\label{Fig:simulations_known}}
\end{figure}

The analysis is performed as follows: for each dataset the posterior probabilities are computed for each observation
\blue{using the true parameters of the model}. The procedure \eqref{Equ : OptimalRule-FDR-KinfP} is then applied to estimate the threshold $\lambda$ in order to control the MFDR at nominal level $\alpha=0.05$. Observations are classified using either the MAP rule, the thresholded rule (with $\alpha=0.05$) or the estimated optimal rule (i.e. the optimal rule applied with estimated threshold $\widehat{\lambda}$). For all rules the actual $MFDR$ and $MFNR$ are evaluated by comparing the predicted and true labels of the classified observations. Results are displayed in Figure \ref{Fig:simulations_known} for the case where all classes are of interest. The case where only classes 1 and 3 are of interest is provided in Appendix D. %\ref{Subsect: SupplMat_Simulations}.

\begin{figure}
  \centering
\includegraphics[scale=0.5]{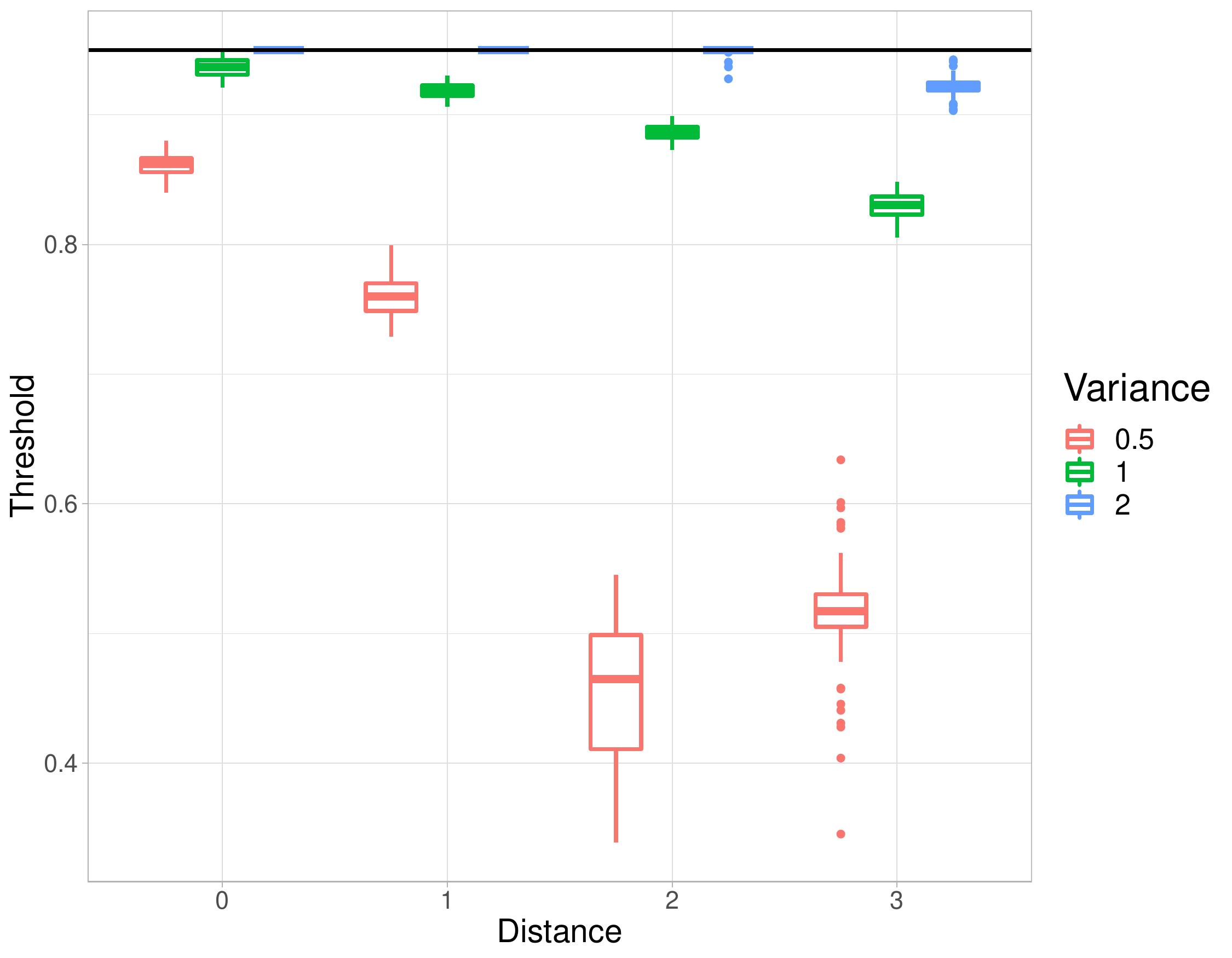}
 \bigskip
 \bigskip
 \caption{Boxplots of the estimated threshold for the optimal rule. Colors correspond to levels of the variance $\sigma^2$. The horizontal line indicates the threshold 0.95 used in the thresholded rule. This figure appears in color in the electronic version of this article.\label{Fig: Lambdas}}
\end{figure}

In the case where $\sigma^2=0.5$ one can observe that all classification rules achieve a low MFDR - that may still be much higher than 0.05 in the case of the MAP rule. Both the thresholded and the optimal rule efficiently achieve control of the MFDR at the nominal level, but with significant differences in terms of MFNR. As illustrated in Figure \ref{Fig: Lambdas}, the estimated threshold for the optimal rule may be much lower (below 0.5 in some configurations) than the $1-\alpha = 0.95$ value used in the thresholded rule. As a consequence, the MFNR is twice higher for the thresholded rule than for the optimal rule, illustrating how conservative the thresholded rule can be.

In the case where $\sigma^2=2$ the classification task becomes quite difficult, and the MAP rule yields MFDR that are higher than 50\%. In such configurations both the thresholded and the optimal rules do not classify any observation in most cases, since the highest posterior probability observed (in a class of interest) is lower than 0.95.

In conclusion, whatever the configuration, the optimal rule $\psi_{\widehat{R}^*}$ with estimated threshold $\widehat{\lambda}$ controls the misclassification rate as requested, whereas the thresholded classification rule is more conservative. When the classification task is easy or intermediate the MFNR of the optimal rule can be much smaller than the MFNR of the thresholded classification rule. Results are unchanged if the prior proportions of the mixture vary (not shown), and/or if only a subset of the three classes are of interest (see Appendix D).

\subsection{\blue{Case 2: posterior probabilities are inferred}}

\begin{figure}
\includegraphics[scale=0.7]{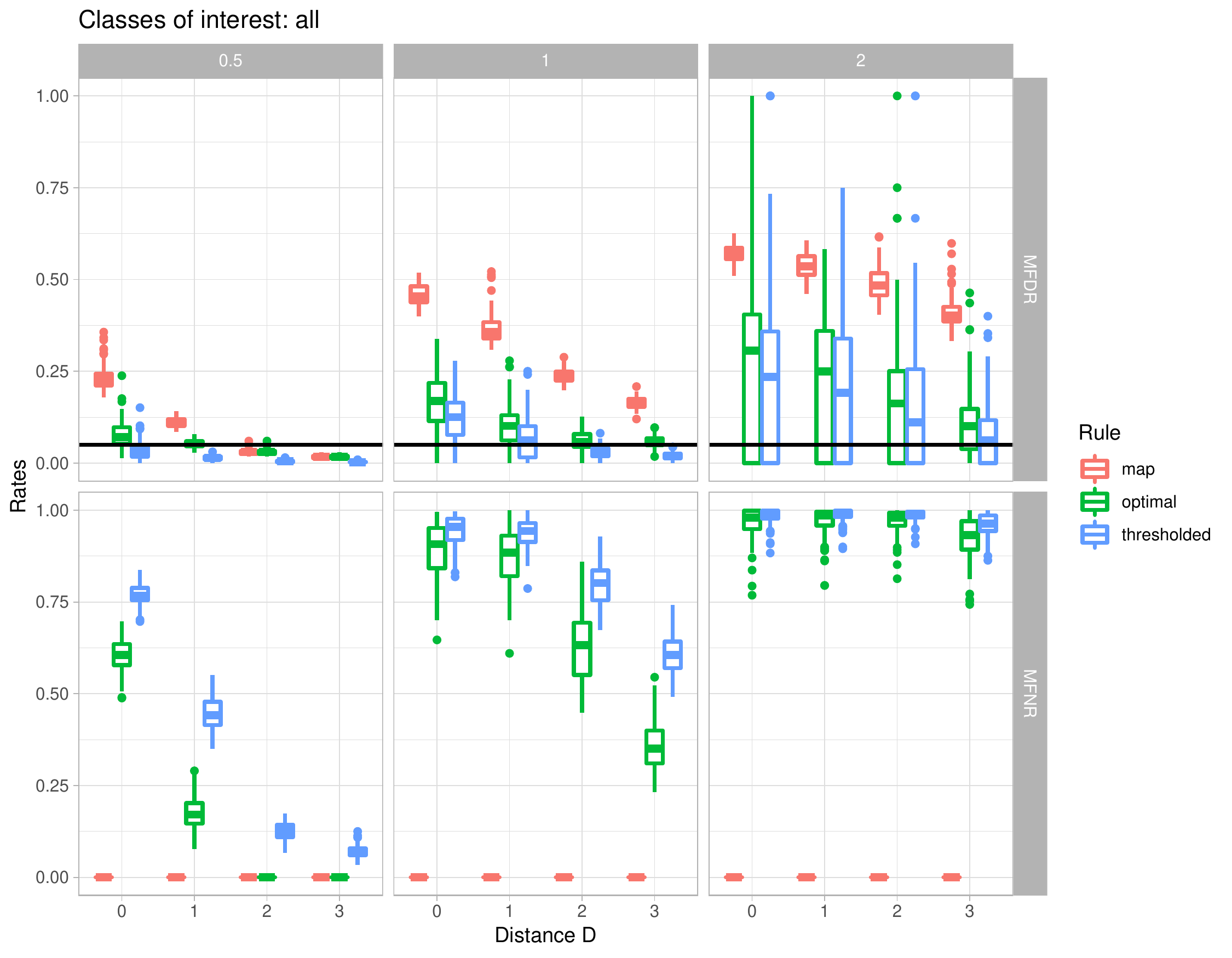}
\bigskip
\bigskip
\caption{Same figure as Figure \ref{Fig:simulations_known}, except that posterior probabilities are now estimated. This figure appears in color in the electronic version of this article. \label{Fig:simulations_est}}
\end{figure}

So far we assumed that the true parameters of the mixture are known, but in practice these parameters are usually estimated from the data at hand.
To evaluate the impact of parameter estimation on misclassification rate, we performed the same analysis as in the previous section, except that parameters are now supposed to be unknown and are estimated using the \blue{\texttt{mclust} R package \citep{Scrucca16}}, fixing the number of classes to its true value (i.e. 3). Results are displayed in Figure \ref{Fig:simulations_est}.

When the classification problem is tractable (low or moderate values of $\sigma^2$ and/or high values of $D$) both the thresholded and the optimal rules efficiently control the MFDR. As in the previous scenario, a significant gain in terms of MFDR is observed when using the optimal rule rather the thresholded rule.
When the classification problem becomes too difficult (i.e. classes strongly overlap) the estimated posterior probabilities get inaccurate, and any classification rule based on these quantities becomes irrelevant. Although neither the thresholded nor the optimal rule control the MFDR at the nominal level, it is still worth to use one of these classification rules rather than the MAP rule. Considering the case where only a subset of classes is of interest leads to similar conclusions (see Figure \ref{Fig:simulations_est_ClInt} in Appendix D).

\blue{The performance of the optimal rule depends on the accuracy of the posterior probability estimates. In order to explore the robustness of our method under model misspecification that could possibly lead to biased posterior probabilities estimates, we generated datasets according to a mixture of three Student bidimensional distributions with the same centers and covariance matrices as in the main Gaussian simulation scheme of ``intermediate'' difficulty ($\sigma=1$). We then estimated the posterior probabilities using the EM algorithm for Gaussian mixtures implemented in \texttt{mclust}. Results are shown in Appendix F. As expected, for higher degrees of freedom, the error rates closely match the ones in the central column ($\sigma=1$) of Figure \ref{Fig:simulations_est}. For lower degrees of freedom, performance breaks down in the ``difficult'' configurations given by low values of $D$. For more tractable configurations, MFDR control is generally achieved and we do not observe notably differences in terms of MFNR with respect to the Gaussian simulations.}

\subsection{Application to transcriptomic data}
% description des données

We consider the unsupervised classification problem described in \cite{Berard2011}. A methylation experiment was performed to compare the methylation profiles of two organs (leaf and seed) of Arabidopsis thaliana. The methylation profiles were measured on a same plant using a tiling array technology. The genomic sequence of Arabidopsis thaliana is represented on the array by approximately $5\times10^5$ probes, covering both genic and intergenic regions. For each probe, the methylation signal is measured in the two organs. From a statistical point of view the sampled population is the population of probes, each of them being described by a bivariate signal (hybridization in leaf and seed), and the goal of the analysis is to identify differentially methylated probes, i.e. probes whose methylation signals in leaf and seed differ. In the following only probes corresponding to chromosome 4 (107,199 probes) are considered. A more thorough description of the data can be found in \cite{Berard2011}.

In \cite{Berard2011} a constrained Gaussian bivariate mixture model was fitted to the data. \blue{Comparing two samples requires distinguishing four different classes of probes that can be biologically interpreted as follows : a class of probes with low methylation signals (1), a class of probes exhibiting a similar intensity level in both organs (2), a class of probes with lower methylation intensities in leaf compared to seed (3), and a symmetric class of probes with higher intensities in leaf compared to seed (4)} (see Appendix E for a graphical representation of the model, and the original article for technical details). \blue{Note that in the initial article the number of classes of the mixture was directly deduced from the biological comparison to be performed}.  Based on this four component mixture model, posterior probabilities to belong to each class were computed for each probe. In the original article the MAP rule was applied to infer class memberships, and probes classified into one of the two classes of interest (under and over-methylated classes) were identified and further investigated. Starting from the same posterior probabilities as in the original publication, we performed probe classification into the classes of interest using the optimal classification rule defined in section \ref{Sect : Shape}. Here $P=4$ and $K=2$, and the MFDR is controlled at nominal level $\alpha=0.1$. Among the 15,801 probes initially classified as under or over-methylated, 13,065 are classified by the resulting MFDR classification rule, yielding an MFNR estimated at 0.0624.

\begin{figure}
\begin{center}
\begin{tabular}{cc}
\includegraphics[width=5cm]{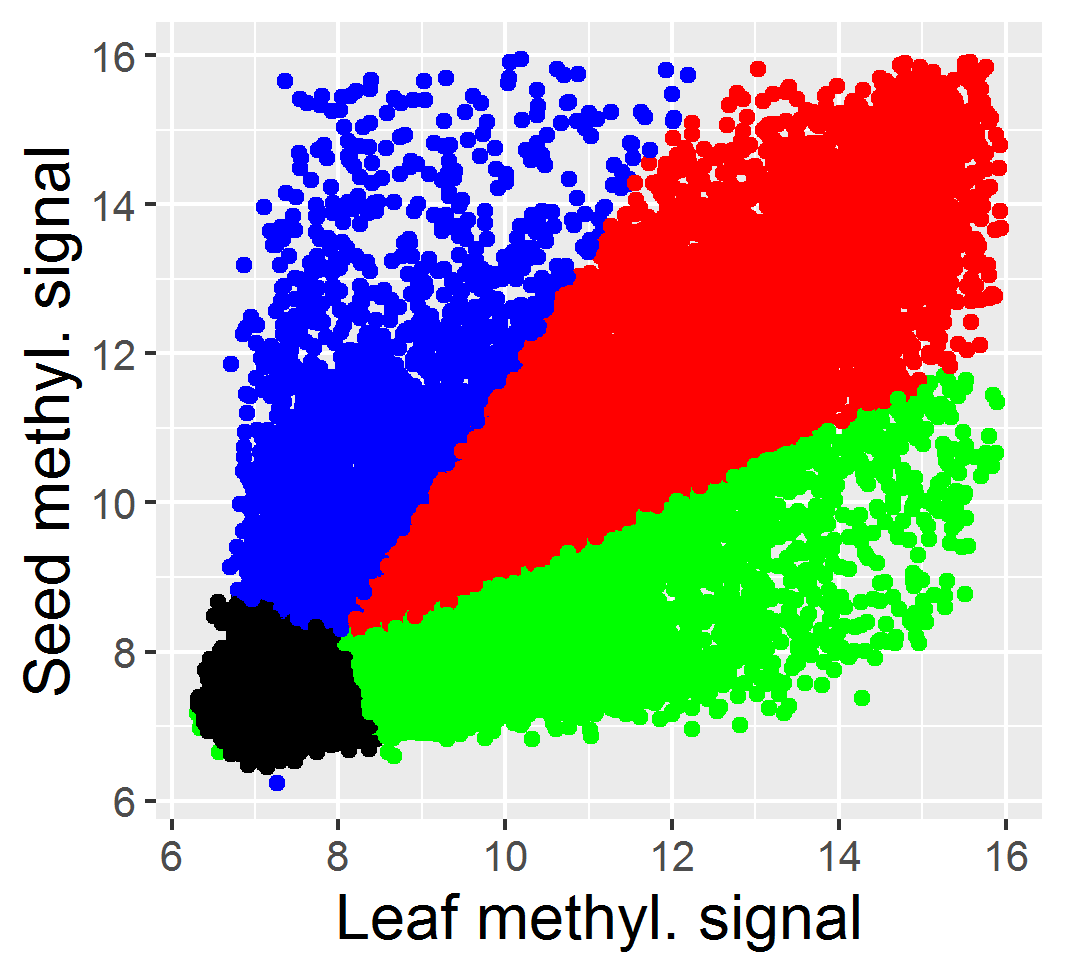} &
\includegraphics[width=5cm]{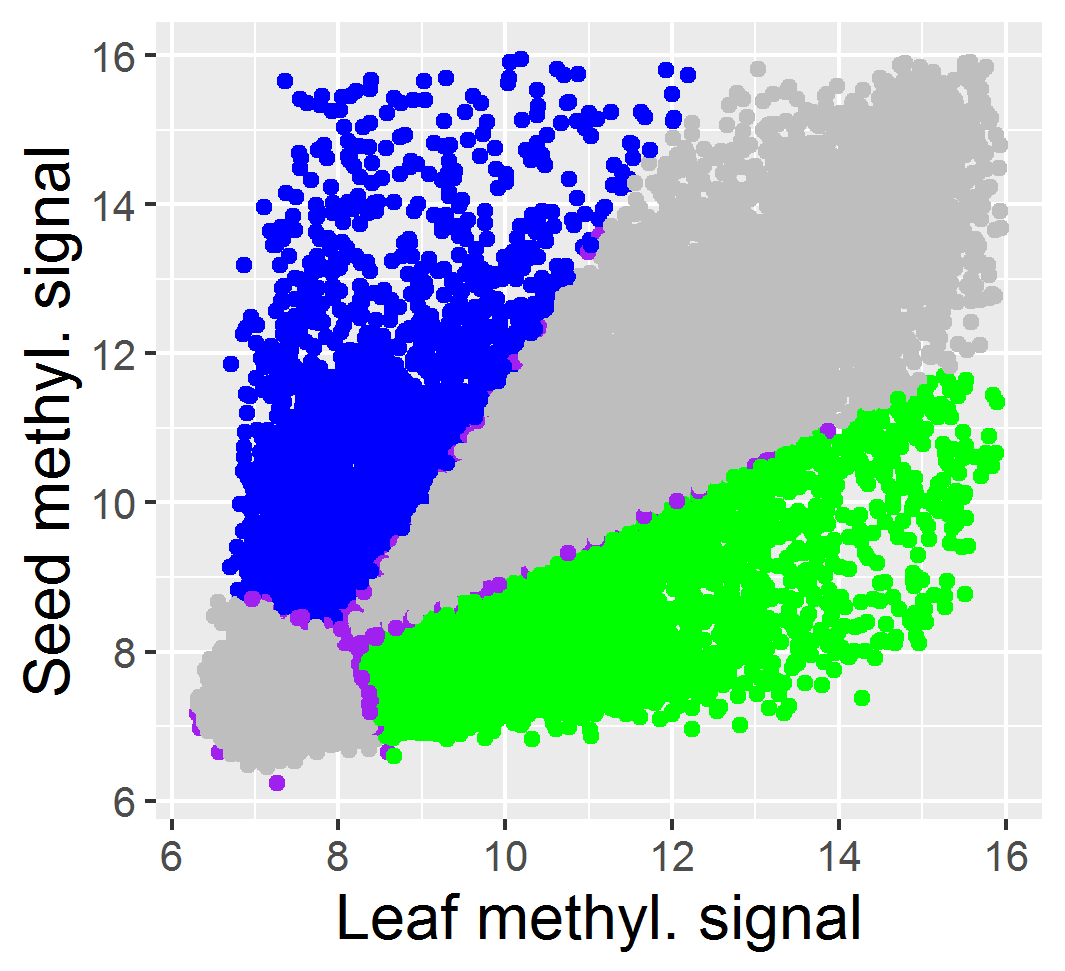}
\includegraphics[width=5cm]{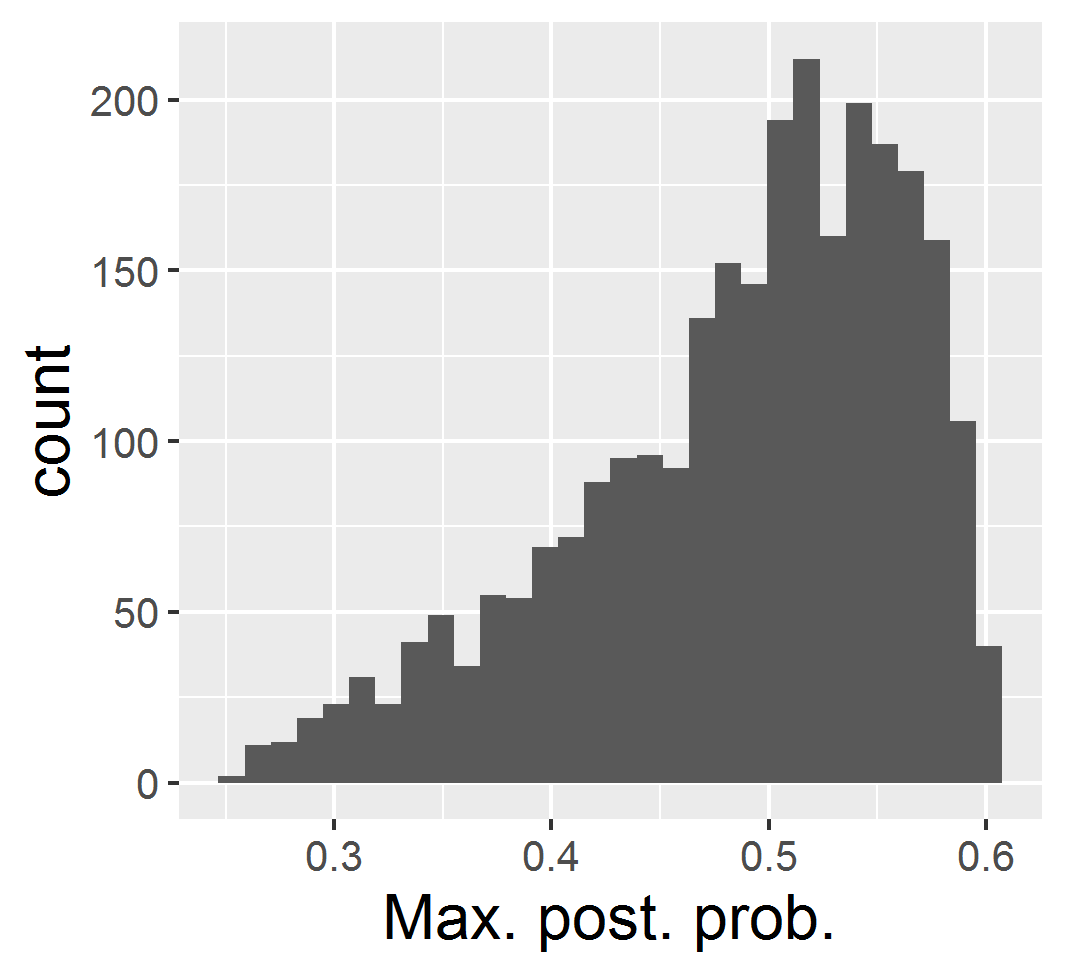}
\end{tabular}
\end{center}
\bigskip
\bigskip
\caption{\textbf{Left:} Probes of chromosome 4 coloured according to the MAP classification rule. \textbf{Center:} Probes of chromosome 4 coloured according to the optimal classification rule. \textbf{Right:} Histogram of the maximum posterior probabilities for purple probes classified by the MAP rule but not by the optimal rule. This figure appears in color in the electronic version of this article.\label{Fig: ClassifiedMethylationData}}
\end{figure}

The different classification rules are illustrated in Figure \ref{Fig: ClassifiedMethylationData}. The left panel represents the MAP rule used in the initial analysis (black = non-methylated, red = identically methylated, blue = under-methylated in leaf, green = over-methylated in leaf). The center panel corresponds to the optimal rule. Here the same colouring of the points is applied, with grey points corresponding to probes unclassified by the optimal rule and purple points corresponding to probes initially classified as over or under-methylated by the MAP rule but unclassified by the optimal rule.

As expected, all purple points are positioned on the boundaries between the two classes of interest and the two other classes. Although the posterior probability to be either over-methylated or under-methylated is higher than any other one for these probes, the actual value of the maximal posterior probability may be quite low, ranging between 0.25 and 0.6 (see Figure \ref{Fig: ClassifiedMethylationData}, right). In contrast, all probes classified by the optimal rule have posterior probabilities higher than 0.57. One can also notice that the optimal classification rule discards some obvious spurious classifications, such as the ones observed for very low values of the methylation signal, that were probably due to the constrained shapes of the covariance matrices of the adjusted model. Such points are discarded due to their low posterior probabilities.

In order to gain some additional insight regarding the boundary probes, we investigated to which extent the classification of a given probe is consistent with the ones of the adjacent (left and right) probes\footnote{Although not accounted for in the initial publication, the information about the genomic position of the probes is available.}. For a given probe, the classification consistency score counts the number of adjacent probes sharing the same classification status. This score takes value 0, 1 or 2, with 2 indicating a high classification consistency between the probe and its two neighbours.

Table \ref{Tab: ClassifVsConsistency} displays the distribution of probes in terms of class assignment and classification consistency, for the MAP rule (left) and the optimal rule (right). Focusing on class ``over-methylated'' one notices that the proportion of probes with score 1 or 2 is higher when using the optimal classification rule rather than the MAP classification rule. This illustrates the fact that most of the probes discarded by the optimal rule (corresponding to the purple points of Figure \ref{Fig: ClassifiedMethylationData}, center) are isolated probes exhibiting weak evidence for methylation. One can conclude that applying the optimal rule yields a more robust set of candidate probes by shaving inconsistent candidates.

\begin{table}
  \begin{small}
    \centerline{
\begin{tabular}{c}
% latex table generated in R 3.5.1 by xtable 1.8-2 package
% Wed Mar 13 18:15:03 2019
\begin{tabular}{rlll}
  \hline
 & 0 & 1 & 2 \\ 
  \hline
Ident. methyl. & 3236 (0.28) & 4374 (0.38) & 4019 (0.35) \\ 
  Non-methyl. & 3159 (0.04) & 18750 (0.24) & 57858 (0.73) \\ 
  Over methyl. & 3469 (0.61) & 1304 (0.23) & 894 (0.16) \\ 
  Under methyl. & 5850 (0.58) & 2544 (0.25) & 1740 (0.17) \\ 
   \hline
\end{tabular}
 \\
(a) \\
% latex table generated in R 3.5.1 by xtable 1.8-2 package
% Wed Mar 13 18:15:03 2019
\begin{tabular}{rlll}
  \hline
 & 0 & 1 & 2 \\ 
  \hline
Not classified & 8514 (0.09) & 23650 (0.25) & 61968 (0.66) \\ 
  Over methyl. & 2650 (0.57) & 1150 (0.25) & 869 (0.19) \\ 
  Under methyl. & 4550 (0.54) & 2172 (0.26) & 1674 (0.2) \\ 
   \hline
\end{tabular}
 \\
 (b)
\end{tabular}\\}
\end{small}
%\vspace{1.5cm}
\caption{\textbf{(a)} Distribution of the probe counts in terms of MAP classification (rows) and classification consistency with neighbours (columns). The four classes correspond to ``non-methylated'' (1), ``identically methylated'' (2), ``under-methylated in leaf'' (3) and ``over-methylated in leaf'' (4). The classification consistency score counts the number of adjacent probes sharing the same classification status as the current probe. Numbers in brackets correspond to per row fractions. \textbf{(b)} Same distribution for the optimal classification. \label{Tab: ClassifVsConsistency}}
%Classification "0" corresponds to unclassified probes.
\end{table}

\section{Discussion \label{Sect : Discussion}}

%We propose methods to build optimal classification rules in the context of finite mixture models, where optimality is meant as the ability to classify as many observations as possible while controlling the rate of misclassification at a given threshold. In particular, for any classification rule, we define two possible misclassification rates (the Multiclass False Discovery Rate and the Multiclass Neyman-Pearson Rate) mimicking the type I error rate in statistical test theory and one error rate (the Multiclass False Negative Rate) inspired by the type II error rate. These definitions make it possible to deal with the situation in which only a subset of the classes are of interest. Finding the optimal classification rule is equivalent to searching for a classification region where to apply the MAP rule restricted to the classes of interest. We show the shape of this region according to the misclassification rate one wants to control and whether all classes, or only a subset, are of interest. We then provide a heuristic to build such regions. The classification rule that controls the MFDR is less conservative than the thresholded MAP rule and achieves better MFDR control than both the MAP and the thresholded MAP rules. We provide a simulation-based study illustrating this claim and apply our method to a real dataset, where we illustrate the advantage of considering our rule.

\blue{The methodology presented here builds on two central ideas: i) in many unsupervised settings only a subset of the classes is of interest for the practitioner, and ii) one would like to provide some guarantees (in terms of error rates) regarding the classification of observations into these specific classes of interest. To this end, classes of interest must be identified beforehand. In some contexts this identification is straightforward. In the methylation application of Section~\ref{Sect : Applications} both the \emph{a priori} number of classes $P$ and the number of classes of interest $K$ are known, a situation that may arise whenever the application context corresponds to a differential analysis setting where two or more conditions are compared. Such examples correspond to the ideal application cases of MFDR control procedures. However, the procedure may also be directly applied to other contexts where $P$ and/or $K$ is unknown. Two examples of such applications are }
\blue{
\begin{itemize}
\item cases where $P$ is unknown but there exists a clear $H_0$ class with known distribution, and the goal is to identify observations that do not belong to the $H_0$ class; here both $P$ and $K$ are initially unknown, and $K$ can be set to $P-1$ once the fitting and selection of the model is performed. Note that in this context and for a fixed nominal level of MFDR, the MFNR could reach different values for different choices of $P$ (hence $K$). In particular, one could expect the MFNR to increase with higher values of $P$. Fixing a maximum level for the MFNR could then guide the model selection, by e.g. choosing $\hat P$ as the largest $P$ satisfying the constraint - a larger $P$ leading to a finer granularity of the classification.
\item cases where both $P$ and $K$ are unknown but prior information about  some observations belonging to a class of interest is available. For instance, in a genomics context one may identify classes of genes based on their expression profiles, then i) identify classes where some genes have a known biological function, and ii) apply the MFDR control procedure to classes corresponding to these functions.
\end{itemize}}
\blue{
In such application cases where either $P$ or $K$ are unknown, the MFDR procedure may be impacted by the model selection, something that was not investigated here. To what extend this impact will affect the procedure is difficult to quantify and will directly depend on the quality of the a priori knowledge one has at hand. For instance in the genomics application mentioned above, applying the MFDR procedure to a well-characterized class of genes may be highly efficient even if the number of classes $P$ is poorly estimated or if some (other) classes are poorly fitted by the inferred model. }\\

While the present paper focused on finding optimal classification rules in an unsupervised framework, a same motivation may exist in the context of (semi-) supervised classification. From a practical point of view, since the optimal classification rules derived in this article only depend on posterior probabilities, they can be applied to any statistical method that yields such probabilities, e.g. logistic regression or discriminant analysis, and could also be extended to methods for which pseudo posterior probabilities can be obtained (see \citealp{Tao05}, \citealp{Grandvalet06} and references therein). 

%From a theoretical point of view, the problem of providing partial rules has already been considered through the use of classification rules with a reject option \citep[see][]{Herbei06,Pillai13}; the setting there is that the reject option has a fixed cost (less costly than a misclassification error), a criterion different from what we have considered here. Alternatively, the problem of controlling a specific error rate related to statistical testing criteria has also been considered in, e.g. \cite{Scott05} and \cite{Scott09} but only in the context of binary classification and without abstention option. These former contributions could be connected to the present work to derive optimal classification rules (in the Neyman Pearson or FDR sense) in the context of multiclass supervised classification.

Future work also includes the definition and derivation of optimal restricted classification rules in contexts where the labels are not assumed to be independent, such as  hidden Markov models and/or latent variables models for network data \citep[stochastic block model and latent block model, see][]{Matias14} where existing results are restricted to the binary classification case \citep{Sun09}. \blue{Additionally the consistency of the MFDR estimates presented here could be investigated using techniques similar to the ones developed in \cite{denis2020}, such that their theoretical results could be extended to the case where $P>2$ and only some classes are of interest. }

\section*{Acknowledgments}
G. Blanchard acknowledges support from Agence Nationale de la Recherche (ANR) via the project ANR-19-CHIA-0021-01 (BiSCottE), and
the project ANR-16-CE40-0019 (SansSouci); and
from  the Franco-German University through the binational Doktorandenkolleg CDFA 01-18.
GQE and IPS2 benefit from the support of the LabEx Saclay Plant Sciences-SPS (ANR-17-EUR-0007).

\bibliographystyle{DeGruyter}
\bibliography{Biblio}

\begin{thebibliography}{27}
\newcommand{\enquote}[1]{``#1''}
\providecommand{\natexlab}[1]{#1}
\providecommand{\url}[1]{\texttt{#1}}
\providecommand{\urlprefix}{URL }

\bibitem[{Bartlett and Wegkamp(2008)}]{bartlett2008}
Bartlett, P. and M.~Wegkamp (2008): \enquote{Classification with a reject
  option using a hinge loss.} \emph{Journal of Machine Learning Research}, 9.

\bibitem[{B{\'e}rard et~al.(2011)B{\'e}rard, Martin-Magniette, Brunaud,
  Aubourg, and Robin}]{Berard2011}
B{\'e}rard, C., M.-L. Martin-Magniette, V.~Brunaud, S.~Aubourg, and S.~Robin
  (2011): \enquote{Unsupervised classification for tiling arrays: Chip-chip and
  transcriptome,} \emph{Statistical applications in genetics and molecular
  biology}, 10.

\bibitem[{Chow(1970)}]{Chow1970}
Chow, C. (1970): \enquote{On optimum recognition error and reject tradeoff,}
  \emph{{IEEE} Transactions on Information Theory}, 16, 41--46.

\bibitem[{Denis and Hebiri(2020)}]{denis2020}
Denis, C. and M.~Hebiri (2020): \enquote{Consistency of plug-in confidence sets
  for classification in semi-supervised learning,} \emph{Journal of
  Nonparametric Statistics}, 32, 42--72.

\bibitem[{Efron and Tibshirani(2002)}]{EfronTibsh02}
Efron, B. and R.~Tibshirani (2002): \enquote{Empirical bayes methods and false
  discovery rates for microarrays,} \emph{Genetic Epidemiology}, 23, 70--86.

\bibitem[{El-Yaniv and Wiener(2010)}]{elyaniv2010}
El-Yaniv, R. and Y.~Wiener (2010): \enquote{On the foundations of noise-free
  selective classification.} \emph{Journal of Machine Learning Research}, 11.

\bibitem[{Friedman et~al.(2009)Friedman, Hastie, and
  Tibshirani}]{friedman2009elements}
Friedman, J., T.~Hastie, and R.~Tibshirani (2009): \emph{The elements of
  statistical learning: data mining, inference, and prediction}, Springer
  series in statistics New York.

\bibitem[{Grandvalet et~al.(2006)Grandvalet, Mari{\'e}thoz, and
  Bengio}]{Grandvalet06}
Grandvalet, Y., J.~Mari{\'e}thoz, and S.~Bengio (2006): \enquote{A
  probabilistic interpretation of svms with an application to unbalanced
  classification,} in \emph{Advances in Neural Information Processing Systems},
  467--474.

\bibitem[{Grandvalet et~al.(2009)Grandvalet, Rakotomamonjy, Keshet, and
  Canu}]{grandvalet2009}
Grandvalet, Y., A.~Rakotomamonjy, J.~Keshet, and S.~Canu (2009):
  \enquote{Support vector machines with a reject option,} in \emph{Advances in
  Neural Information Processing Systems 21 (NIPS 2008)}, MIT press, 537--544.

\bibitem[{Herbei and Wegkamp(2006)}]{Herbei06}
Herbei, R. and M.~H. Wegkamp (2006): \enquote{Classification with reject
  option,} \emph{Canadian Journal of Statistics}, 34, 709--721.

\bibitem[{Karmakar et~al.(2019)Karmakar, Das, Bhattacharya, Sarkar, and
  Mukhopadhyay}]{Karmakar2019}
Karmakar, B., S.~Das, S.~Bhattacharya, R.~Sarkar, and I.~Mukhopadhyay (2019):
  \enquote{{Tight clustering for large datasets with an application to gene
  expression data},} \emph{Scientific Reports}, 9, 3053.

\bibitem[{Lei(2014)}]{Lei2014}
Lei, J. (2014): \enquote{Classification with confidence,} \emph{Biometrika},
  101, 755--769.

\bibitem[{Matias and Robin(2014)}]{Matias14}
Matias, C. and S.~Robin (2014): \enquote{Modeling heterogeneity in random
  graphs through latent space models: a selective review,} \emph{ESAIM:
  Proceedings and Surveys}, 47, 55--74.

\bibitem[{McLachlan and Peel(2000)}]{McLachPeel00}
McLachlan, G.~J. and D.~Peel (2000): \emph{Finite mixture models}, New York:
  Wiley.

\bibitem[{Neyman and Pearson(1933)}]{neyman1933ix}
Neyman, J. and E.~S. Pearson (1933): \enquote{On the problem of the most
  efficient tests of statistical hypotheses,} \emph{Philosophical Transactions
  of the Royal Society of London. Series A, Containing Papers of a Mathematical
  or Physical Character}, 231, 289--337.

\bibitem[{Pillai et~al.(2013)Pillai, Fumera, and Roli}]{Pillai13}
Pillai, I., G.~Fumera, and F.~Roli (2013): \enquote{Multi-label classification
  with a reject option,} \emph{Pattern Recognition}, 46, 2256--2266.

\bibitem[{Schreuder and Chzhen(2021)}]{schreuder2021}
Schreuder, N. and E.~Chzhen (2021): \enquote{Classification with abstention but
  without disparities,} .

\bibitem[{Scott et~al.(2009)Scott, Bellala, Willett et~al.}]{Scott09}
Scott, C., G.~Bellala, R.~Willett, et~al. (2009): \enquote{The false discovery
  rate for statistical pattern recognition,} \emph{Electronic Journal of
  Statistics}, 3, 651--677.

\bibitem[{Scott and Nowak(2005)}]{Scott05}
Scott, C. and R.~Nowak (2005): \enquote{A {N}eyman-{P}earson approach to
  statistical learning,} \emph{IEEE Transactions on Information Theory}, 51,
  3806--3819.

\bibitem[{Scrucca et~al.(2016)Scrucca, Fop, Murphy, and Raftery}]{Scrucca16}
Scrucca, L., M.~Fop, T.~Murphy, and A.~Raftery (2016): \enquote{{mclust} 5:
  clustering, classification and density estimation using {G}aussian finite
  mixture models,} \emph{The {R} Journal}, 8, 289--317.

\bibitem[{Sun and Cai(2009)}]{Sun09}
Sun, W. and T.~T. Cai (2009): \enquote{Large-scale multiple testing under
  dependence,} \emph{Journal of the Royal Statistical Society: Series B
  (Statistical Methodology)}, 71, 393--424.

\bibitem[{Tao et~al.(2005)Tao, Wu, Wang, and Wang}]{Tao05}
Tao, Q., G.-W. Wu, F.-Y. Wang, and J.~Wang (2005): \enquote{Posterior
  probability support vector machines for unbalanced data,} \emph{IEEE
  Transactions on Neural Networks}, 16, 1561--1573.

\bibitem[{Tong et~al.(2016)Tong, Feng, and Zhao}]{tong2016}
Tong, X., Y.~Feng, and A.~Zhao (2016): \enquote{A survey on neyman-pearson
  classification and suggestions for future research,} \emph{Wiley
  Interdisciplinary Reviews: Computational Statistics}, 8, 64--81.

\bibitem[{Tseng and Wong(2005)}]{Tseng2005}
Tseng, G.~C. and W.~H. Wong (2005): \enquote{Tight clustering: A
  resampling-based approach for identifying stable and tight patterns in data,}
  \emph{Biometrics}, 61, 10--16.

\bibitem[{Wegkamp and Yuan(2011)}]{wegkamp2011}
Wegkamp, M. and M.~Yuan (2011): \enquote{Support vector machines with a reject
  option,} \emph{Bernoulli}, 17, 1368--1385.

\bibitem[{Wiener and El-Yaniv(2015)}]{wiener2015}
Wiener, Y. and R.~El-Yaniv (2015): \enquote{Agnostic pointwise-competitive
  selective classification,} \emph{Journal of Artificial Intelligence
  Research}, 52, 171--201.

\bibitem[{Zhang and Chaudhuri(2014)}]{zhang2014}
Zhang, C. and K.~Chaudhuri (2014): \enquote{Beyond disagreement-based agnostic
  active learning,} in \emph{Advances in Neural Information Processing Systems
  27 (NIPS 2014)}, 442--450.

\end{thebibliography}

\newgeometry{letterpaper,tmargin=2cm,bmargin=2cm,lmargin=2cm,rmargin=2cm}

\section*{Appendix A: Proof of Proposition \ref{Prop:MNPR classification region}}

%\begin{proof}
%Assuming the marginal distribution of $X$ to be without atoms, 

  \blue{We denote $\psi$ the MAP rule amongst class
  1 to $K$, and $\psi_R$ this rule restricted to a region $R$. If $MNPR(\psi_\X)\leq \alpha$, since $MFNR(\psi_R) \geq MFNR(\psi_\X)$ for any region $R$, obviously $R^*=\X$ is a solution.}
  
  \blue{Otherwise, let $R^*$ be a region satisfying the conditions of the
  theorem, and let $R$ be any region} such that the MAP classification rule restricted to $R$ satisfies $MNPR(\psi_{R}) \leq  \alpha = MNPR(\psi_{R^*})$. Let us prove that $MFNR(\psi_{R})\geq MFNR(\psi_{R^*})$. Observing that $R = R \cap (R^*\cup \bar{R^*})$ and $R^* = R^* \cap (R \cup \bar{R})$, one has :
\begin{eqnarray*}
&&\prob{\psi(X) \neq Z , \  X\in R }  \leq  \prob{\psi(X) \neq Z , \  X\in R^* }\\
&\Leftrightarrow& \prob{\psi(X) \neq Z , \  X\in R\cap\bar{R^*} } \leq
\prob{\psi(X) \neq Z , \  X\in R^*\cap\bar{R} }\\
&\Leftrightarrow& \int_{R\cap \bar{R^*}} (1-\tau^*_K(x))f(x)dx \leq
\int_{R^*\cap \bar{R}} (1-\tau^*_K(x))f(x)dx ,
\end{eqnarray*}
where the last equation follows from the application of the MAP rule on $R$ and $R^*$ respectively.

Similarly,
\begin{eqnarray*}
&&P\gp{Z\in\{1,\ldots,K\} , \  X\in \bar{R}} - P\gp{Z\in\{1,\ldots,K\} , \  X\in \bar{R^*}} \geq 0 \\
&\Leftrightarrow& P\gp{Z\in\{1,\ldots,K\} , \  X\in \bar{R}\cap R^*} - P\gp{Z\in\{1,\ldots,K\} , \  X\in
\bar{R^*}\cap  R} \geq 0.
\end{eqnarray*}
Therefore
\begin{eqnarray*}
P\gp{Z\in\{1,\ldots,K\} , \   X\in \bar{R}\cap  R^*} & = & \int_{\bar{R}\cap  R^*}
\gp{\sum_{k=1}^K\tau_k(x)}f(x)dx \\
&\geq& \lambda\int_{\bar{R}\cap  R^*}(1-\tau^*_K(x))f(x)dx\\
&\geq& \lambda\int_{\bar{R^*}\cap  R}(1-\tau^*_K(x))f(x)dx\\
&\geq& \int_{\bar{R^*}\cap  R}\gp{\sum_{k=1}^K\tau_k(x)}f(x)dx \\
&\blue{=}& P\gp{Z\in\{1,\ldots,K\},\ X\in \bar{R^*}\cap  R},
\end{eqnarray*}
where the first and third inequalities above follow from the property of $R^*$ (and thus $\bar{R^*}$) given in equation \eqref{Equ : OptimalRule-NP-KinfP }.

\blue{Assuming that the marginal distribution of $X$
is without atoms, we now establish that there exists
    a region $R^*$ satisfying~\eqref{Equ : OptimalRule-NP-KinfP } for some $\lambda\geq 0$, and satisfying the MNPR constraint with equality,
    provided $MNPR(\psi_\X) >\alpha$. %Recall $\psi(x)$ is the MAP rule, a fixed function,
    Define the event $A=\{\psi(X) \neq Z\}$, and
    $Q$ the measure on the space $\X$ 
    as $Q(R) = P( \{X \in R\} \cap A) = MNPR(\psi_{R})$. Since the
    marginal distribution of $X$ is without atoms, so is $Q$.}

\blue{    Denote $\xi(x)=(\sum_{k=1}^K \tau_k(x))/(1-\tau^*_K(x))$, and define the level sets for any $\lambda\geq 0$:
    \[
        L(\lambda) = \{ x \in \X: \xi(x) \geq \lambda \} ; \qquad  L^>(\lambda) = \{ x \in \X: \xi(x) > \lambda \}.
    \]
    Observe that it holds $L^>(\lambda) \subseteq L(\lambda)$ for any
    $\lambda$, and $L(\lambda) = \bigcap_{\lambda' < \lambda} L^>(\lambda')$, 
    $L^>(\lambda) = \bigcup_{\lambda' > \lambda} L^>(\lambda')$, so that
    \begin{equation}
        \label{eq:encadrement}
    \limsup_{\lambda'\searrow \lambda} Q(L^>(\lambda')) = Q(L^>(\lambda)) \leq Q(L(\lambda)) = \liminf_{\lambda' \nearrow \lambda} Q(L^>(\lambda')).
    \end{equation}
    Define
    \[
    \lambda^* = \sup \{\lambda \geq 0: Q(L^>(\lambda)) \geq \alpha\} \cup \{0\},    \] 
 %   where the set on the right-hand side is non-empty since
 %   we assumed $Q(\X)=MNPR(\psi_\X)>\alpha$.
    By the definition of $\lambda^*$, it holds
    \[
        Q(L^>(\lambda^*)) \leq \alpha \leq  Q(L(\lambda^*));
    \]
    this is true by~\eqref{eq:encadrement} if $\lambda^*>0$,
    and, if $\lambda^*=0$, it holds as well since
    we assumed $Q(L(0)) = Q(\X) = MNPR(\psi_\X)>\alpha$.
    (It can be also checked that it still holds if
    $\lambda^*=\infty$, which can happen in principle when
    $\prob{\xi(X)=\infty}>0$.)
%    Denote $\Delta^* = L(\lambda^*)\setminus L^>(\lambda^*) = A \cap \{ \xi(X) = \lambda^*\}.$
%    , so there exists $B$
    Therefore, since $Q$ is without atoms, there exists a set $R\subseteq \X$ such that
    $L^>(\lambda^*) \subseteq R \subseteq L(\lambda^*))$ with $Q(R) = MNPR(\psi_R) =\alpha$, as required.
  %  Since we assumed that the marginal distribution of $X$ is without atoms, it follows that
  %  the joint distribution of $(X,Z)$ is also without atoms.
    }
%\end{proof}

\section*{Appendix B: Proof of Proposition \ref{Prop:MFDR classification region}}

%\begin{proof}
\blue{If $MFDR(\psi_\X)\leq \alpha$, since $MFNR(\psi_R) \geq MFNR(\psi_\X)$ for any region $R$, obviously $R^*=\X$ is a solution.
For any region $R\subseteq \X$ with $P(X \in R)>0$, it holds that
$MFDR(\psi_R) = \esp{1-\tau^*_K(X)|X\in R}$.
%Therefore, if $\tau^*_K(X)\geq 1-\alpha$ almost surely,
%it holds $MFDR(\psi_R) \leq \alpha$ for any region $R$. In 
%this case, the MFNR is maximized for $R^*=\X$.
Therefore, if $\tau^*_K(X)< 1-\alpha$ almost surely,
it holds $MFDR(\psi_R) > \alpha$ for any region $R$
with $P(X \in R)>0$. Thus, in this case only a region $R$ with null probability (in particular $R^*=\emptyset$) 
satisfies the MFDR constraint since by convention  $MFDR(\Psi_\emptyset)=0$.}

Otherwise, let $R^*$ be a region satisfying~\eqref{Equ : OptimalRule-FDR-KinfP} and
    the MFDR constraint with equality.
Consider $R$ any region satisfying
$$
\prob{\psi(X) \neq Z | \ X\in R } = \prob{ \psi(X) \neq Z | \ X\in R^*} = \alpha.
$$
On the one hand, one has :
\begin{eqnarray*}
\prob{X\in R}-\prob{X\in R^*}&=&\prob{ X\in R\cap
\bar{R}^*}-\prob{X\in R^*\cap  \bar{R}} \ \ .
\end{eqnarray*}
On the other hand,
\begin{eqnarray*}
\prob{\psi(X) \neq Z| \ X\in R } = \alpha &\Leftrightarrow&\prob{\psi(X) \neq Z , \  X\in R } = \alpha\prob{X\in
R}\\
\Rightarrow \alpha\left[\prob{X\in R}-\prob{X\in R^*} \right] &=&
\prob{\psi(X) \neq Z , \  X\in R } - \prob{\psi(X) \neq Z,
\ X\in R^* }  \\
&=& \prob{\psi(X) \neq Z , \  X\in R\cap  \bar{R}^* } -
\prob{\psi(X) \neq Z , \  X\in R^*\cap  \bar{R} } \ \ .
\end{eqnarray*}
Hence
\begin{eqnarray*}
&&\hspace{-1cm}\prob{\psi(X) \neq Z , \  X\in R \cap \bar{R}^* } -
\prob{\psi(X) \neq Z , \  X\in R^*\cap \bar{R} }=\alpha\left[\prob{
X\in R \cap \bar{R}^*}-\prob{X\in
R^*\cap  \bar{R}} \right]\\
&\Rightarrow& \int_{R\small{\cap}
\bar{R}^*}(1-\tau^*_K(x)-\alpha)f(x)dx
- \int_{R^*\small{\cap} \bar{R}}(1-\tau^*_K(x)-\alpha)f(x)dx=0 \\
&\Rightarrow& \lambda\gp{\int_{R\small{\cap}
\bar{R}^*}\gp{\sum_{k=1}^K\tau_k(x)}f(x)dx
- \int_{R^*\small{\cap} \bar{R}}\gp{\sum_{k=1}^K\tau_k(x)}f(x)dx}\leq 0 \\
&\Rightarrow&\prob{Z\in\{1,\ldots,K\},  \ X\in R\small{\cap} \bar{R}^* }-\prob{Z\in\{1,\ldots,K\},
\ X\in R^*\small{\cap} \bar{R} } \leq 0\\
&\Rightarrow&\prob{Z\in\{1,\ldots,K\},  \ X\in \bar{R} }-\prob{Z\in\{1,\ldots,K\}, \ X\in
\bar{R^*} } \geq 0.
\end{eqnarray*}
\blue{Analogously to the proof of Proposition~\ref{Prop:MNPR classification region}, we establish the existence of $R^*$
satisfying the required conditions, provided we exclude the
edge cases, i.e. we assume that $MFDR(\psi_\X) >\alpha$,
and also $\prob{\tau^*_K(X)\geq 1-\alpha}>0$.}

\blue{Recall $MFDR(\psi_R) = \esp{1-\tau^*_K(X)|X\in R}$ (provided $\prob{X \in R}>0$), so $MFDR(\psi_R) = \alpha$
is equivalent to $\esp{ (1-\tau^*_K(X) -\alpha) \ind{X \in R}}=0$. Let $Q$ be the signed measure on the space $\X$ 
defined as
$Q(R) = \esp{(1-\tau^*_K(X) -\alpha) \ind{X \in R}}$.}

\blue{Denote $\xi(x)=(1-\tau^*_K(x)-\alpha)/(\sum_{k=1}^K \tau_k(x))$, and define the level sets for any $\lambda\geq 0$:
    \[
        L(\lambda) = \{ x \in \X: \xi(x) \leq \lambda \} ; \qquad  L^<(\lambda) = \{ x \in \X: \xi(x) < \lambda \}.
    \]
    Observe that it holds $L^<(\lambda) \subseteq L(\lambda)$ for any
    $\lambda$, and $L(\lambda) = \bigcap_{\lambda' > \lambda} L^<(\lambda')$, 
    $L^<(\lambda) = \bigcup_{\lambda' < \lambda} L^<(\lambda')$, so that
    \begin{equation}
        \label{eq:encadrement2}
    \limsup_{\lambda'\nearrow \lambda} Q(L^<(\lambda')) = Q(L^<(\lambda)) \leq Q(L(\lambda)) = \liminf_{\lambda' \searrow \lambda} Q(L^<(\lambda')).
    \end{equation}
    Define
    \[
    \lambda^* = \inf \{\lambda \geq 0: Q(L^<(\lambda)) \geq 0\};    \] 
    we take $\lambda^*=\infty$ if the above set is empty.
 %   where the set on the right-hand side is non-empty since
 %   we assumed $Q(\X)=MNPR(\psi_\X)>\alpha$.
    By the definition of $\lambda^*$, it holds
    \[
        Q(L^<(\lambda^*)) \leq 0 \leq  Q(L(\lambda^*));
    \]
    this is true by~\eqref{eq:encadrement2} if $0<\lambda^*<\infty$;
    if $\lambda^*=\infty$, it holds as well because $Q(L(\infty)) = Q(\X) >0$ since $MFDR(\psi_\X)>\alpha$;
    if $\lambda^*=0$, it still holds because $Q(L^<(0))\leq 0$ in general.
%    Denote $\Delta^* = L(\lambda^*)\setminus L^>(\lambda^*) = A \cap \{ \xi(X) = \lambda^*\}.$
%    , so there exists $B$
    Therefore, since $Q$ is without atoms, there exists a set $R\subseteq \X$ such that
    $L^>(\lambda^*) \subseteq R \subseteq L(\lambda^*)$ with $Q(R) = 0 $.
    This will imply $MFDR(\psi_R) =\alpha$, as required, provided
    $\prob{X\in R}>0$, which we still have to check to finish the proof.
  %  Since we assumed that the marginal distribution of $X$ is without atoms, it follows that
  %  the joint distribution of $(X,Z)$ is also without atoms.$
}

\blue{Recall we assumed $\prob{\tau^*_K(X)\geq 1-\alpha)}>0$.
Since  $\{x \in \X: \tau^*_K(X)\geq 1-\alpha)\}=L(0)$, this means 
$\prob{X \in L(0)} >0$. If $\lambda^*>0$, it holds $L(0) \subseteq R$ 
and we are finished. In the special case
$\lambda^*=0$, it holds both $Q(L(0))\leq 0$ by definition of $L(0)$,
and $Q(L(0)) \geq 0$ by definition of $\lambda^*$. Hence $Q(L(0))=0$
in this case, and we can take $R=L(0)$.}
%\end{proof}

\section*{Appendix C: Illustration of different configurations of the simulation study}

\begin{figure}[ht!]
  \centering
 \begin{tabular}{ccc}
 \includegraphics[scale=0.25]{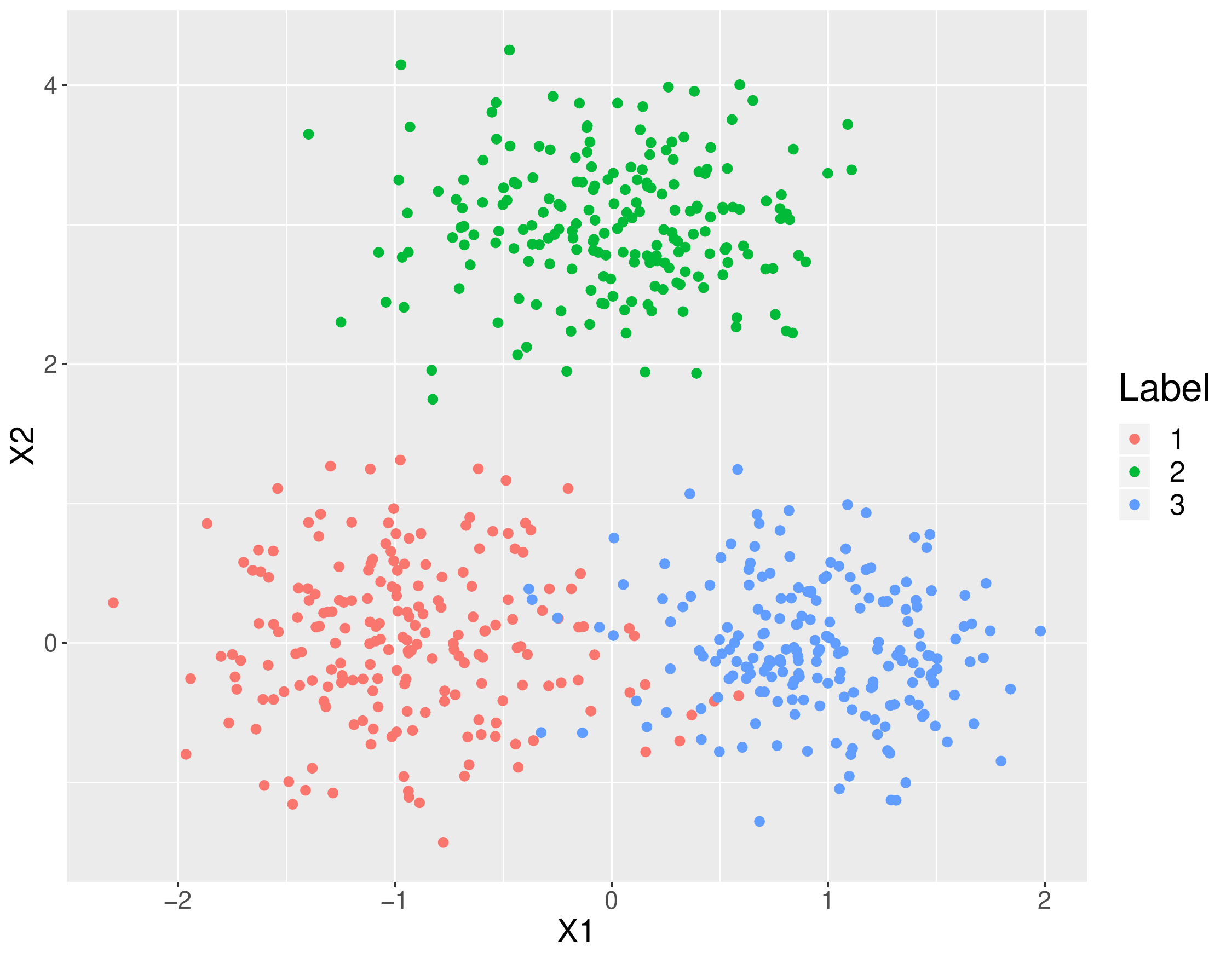}&
 \includegraphics[scale=0.25]{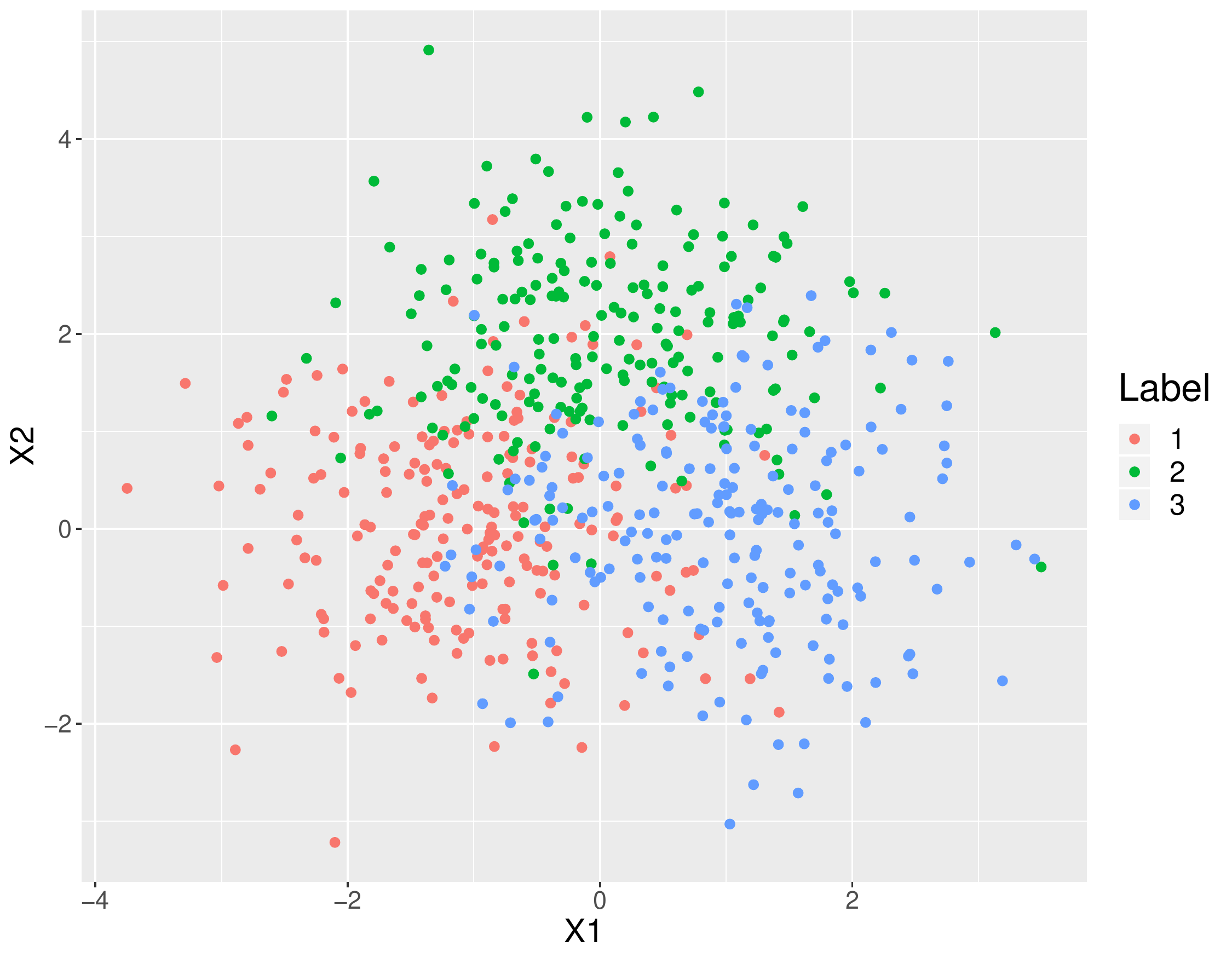}
 \includegraphics[scale=0.25]{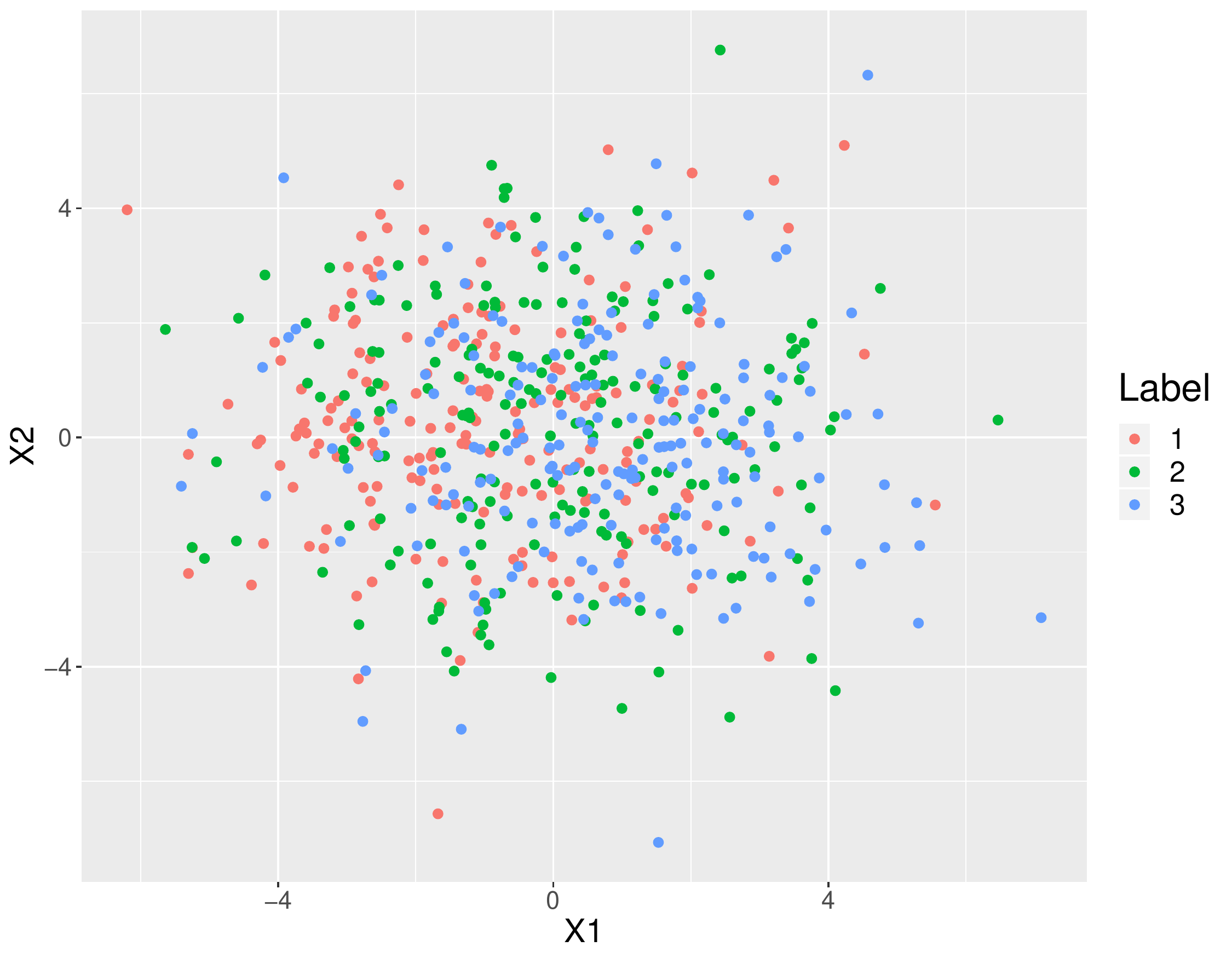}
 \end{tabular}
 \bigskip
 \bigskip
 \caption{Three examples of simulated data, with different parameter values: an easy case (left) corresponding to $D=3$ and $\sigma^2=0.5$, and intermediate case (center) corresponding to $D=2$ and $\sigma^2=1$ and a difficult case (right) corresponding to  $D=0$ and $\sigma^2=2$. Colors correspond to class labels. \label{fig:data}}
\end{figure}

\section*{Appendix D: Analysis of simulated data when only some classes are of interest.} \label{Subsect: SupplMat_Simulations}
\begin{figure}[ht!]
\includegraphics[scale=0.8]{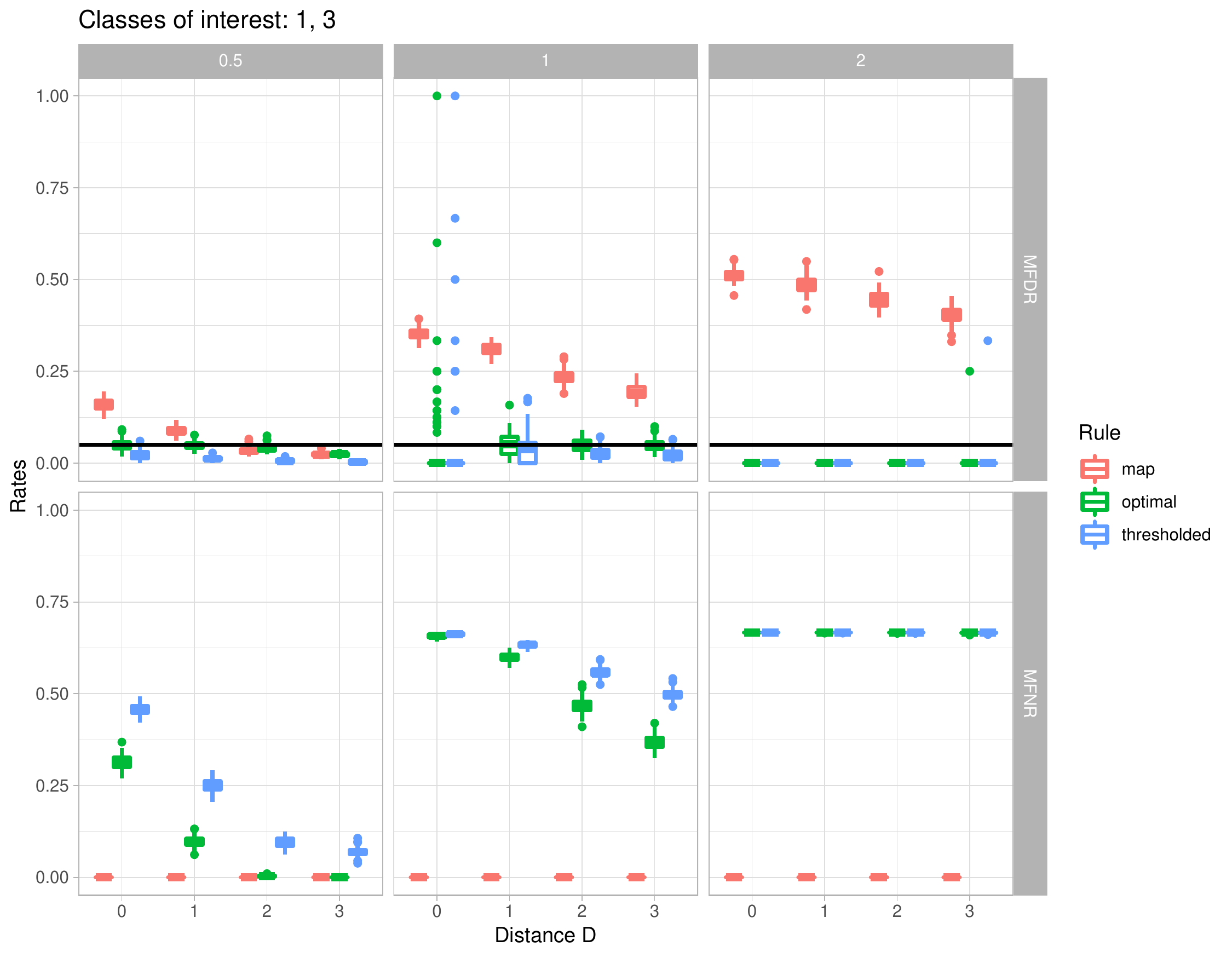}
\caption{ Performances of the MAP, optimal and $1-\alpha$ thresholded classification rules in terms of estimated and true MFDR and MFNR. All rules are based on the true posterior probabilities. Only classes 1 and 3 are assumed to be of interest. \label{Fig:simulations_known_ClInt}}
\end{figure}

\begin{figure}[ht!]
\includegraphics[scale=0.8]{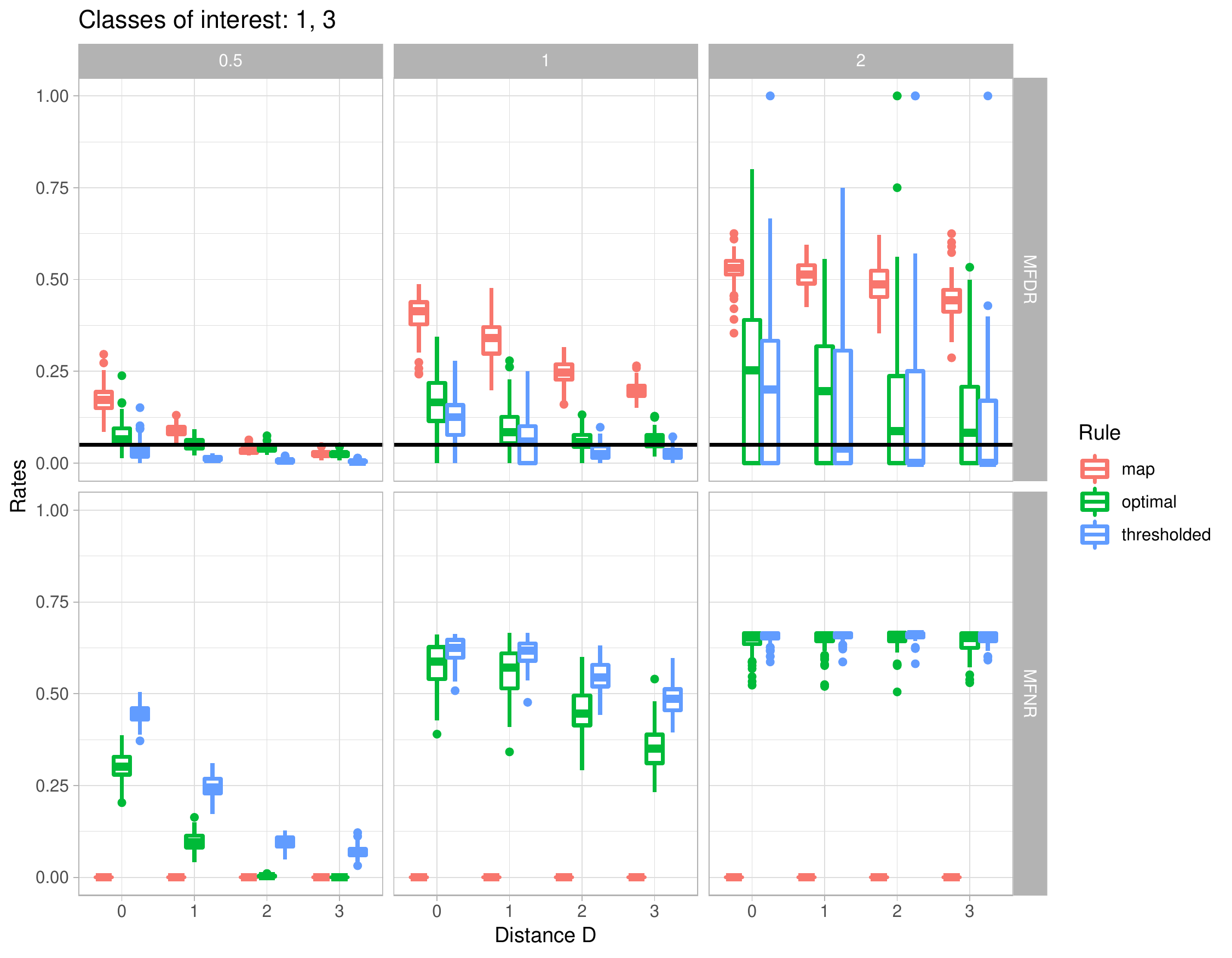}
\caption{ Same Figure as \ref{Fig:simulations_known_ClInt}, except that posterior probabilities are inferred from the data. \label{Fig:simulations_est_ClInt}}
\end{figure}

Here the same simulation setting is considered as in the main article, the only difference being that it is now assumed that only classes 1 and 3 are of interest. Figure \ref{Fig:simulations_known_ClInt} displays the performance of the different classification rules when the true parameters of the model are known. Figure \ref{Fig:simulations_est_ClInt} displays the same result when the parameters of the model are inferred from the data.

\vspace*{-8pt}
\clearpage

\section*{Appendix E: Illustration of the dataset and model used in Berard et al. (2011).}
\begin{figure}[ht!]
\begin{center}
\begin{tabular}{cc}
\includegraphics[width=7.5cm]{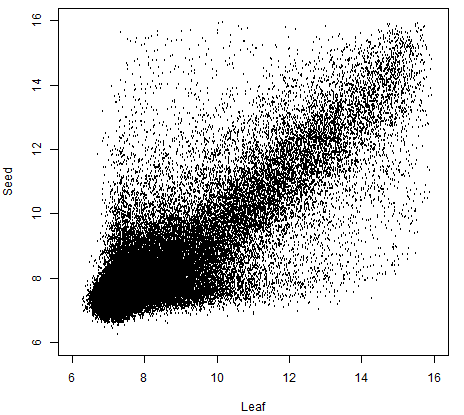} &
\includegraphics[width=7.5cm]{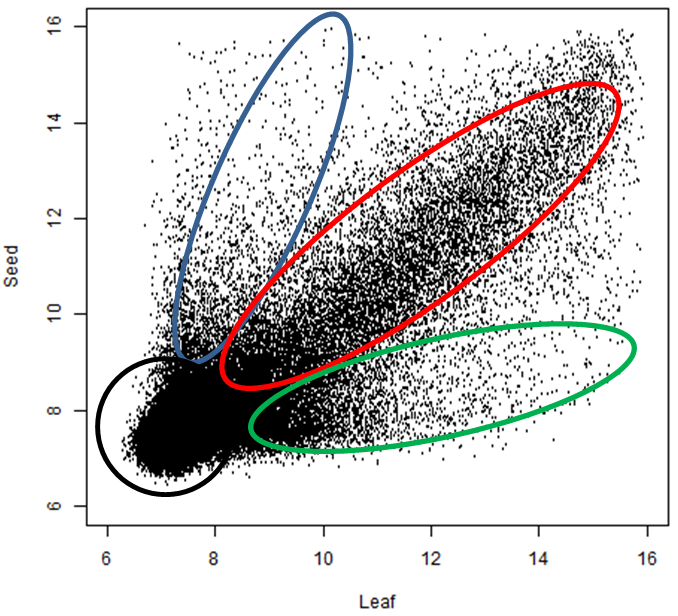}
\end{tabular}
\end{center}
\bigskip
\bigskip
\caption{\textbf{Left:} Methylation data for chromosome 4. Each point corresponds to a probe, represented by its methylation levels in the leaf (x-axis) and seed (y-axis) \textbf{Right:} Same graph with the four ellipses corresponding to the four Gaussian bivariate components. \label{Fig: MethylationData}}
\end{figure}

Figure \ref{Fig: MethylationData} (left) provides a graphical representation of the dataset. Each point corresponds to a probe, represented by its methylation signal in the leaf (x-axis) and in the seed (y-axis). One can observe four different clouds: the upper (respectively lower) cloud corresponds to probes that are over-methylated (resp. under methylated) in the seed compared to the leaf. The two other clouds are positioned on the first bisector and correspond to non methylated probes (signals close to 0 in the two organs) or identically methylated probes. The model is graphically represented on Figure \ref{Fig: MethylationData} (right), where the four ellipses represent the (constrained) shapes of the covariance matrices of the four Gaussian components of the constrained Gaussian bivariate mixture.

\section*{Appendix F: Simulations based on a mixture of Student bivariate distributions}
\label{Subsect: SupplMat_Simulations_Student}

\blue{For this supplementary simulation study, datasets were simulated from a mixture of three bidimensional Student distributions. We recall that given a Gaussian vector $\mathbf y\sim\mathcal{N}(\mathbf0,\boldsymbol{\Sigma})$, a location vector $\boldsymbol{\mu}$ and $u\sim\chi_{\nu}^2$, the vector
$$
\mathbf x = \frac{\mathbf y}{\sqrt{\frac{u}{\nu}}} + \boldsymbol{\mu}
$$
follows a multivariate Student distribution with parameters $\boldsymbol{\Sigma}, \boldsymbol \mu, \nu$. 
For each of the three components in the mixture, we simulated 200 observations with the same locations $\boldsymbol{\mu_1} = (-1,0),\boldsymbol{\mu_2} = (0,D)$ and $\boldsymbol{\mu_3} = (1,0)$ as in our principal Gaussian simulations. We took the scale matrix $\boldsymbol{\Sigma}$ to be the diagonal matrix $\sigma'^2\boldsymbol{I}$ where $\sigma'$ was chosen so that the covariance matrix $\frac{\nu}{\nu-2}\boldsymbol{\Sigma}$ is the same as the one in the Gaussian scenario of ``intermediate'' difficulty, i.e. $\frac{\nu}{\nu-2}\sigma'^2=1$. We considered several degrees of freedom $\nu$ ($5,10,20,50$), and the same values of $D$ as in our Gaussian simulations ($0,1,2,3$). We generated 100 simulations for each configuration.}

\begin{figure}[ht!]
\includegraphics[scale=0.8]{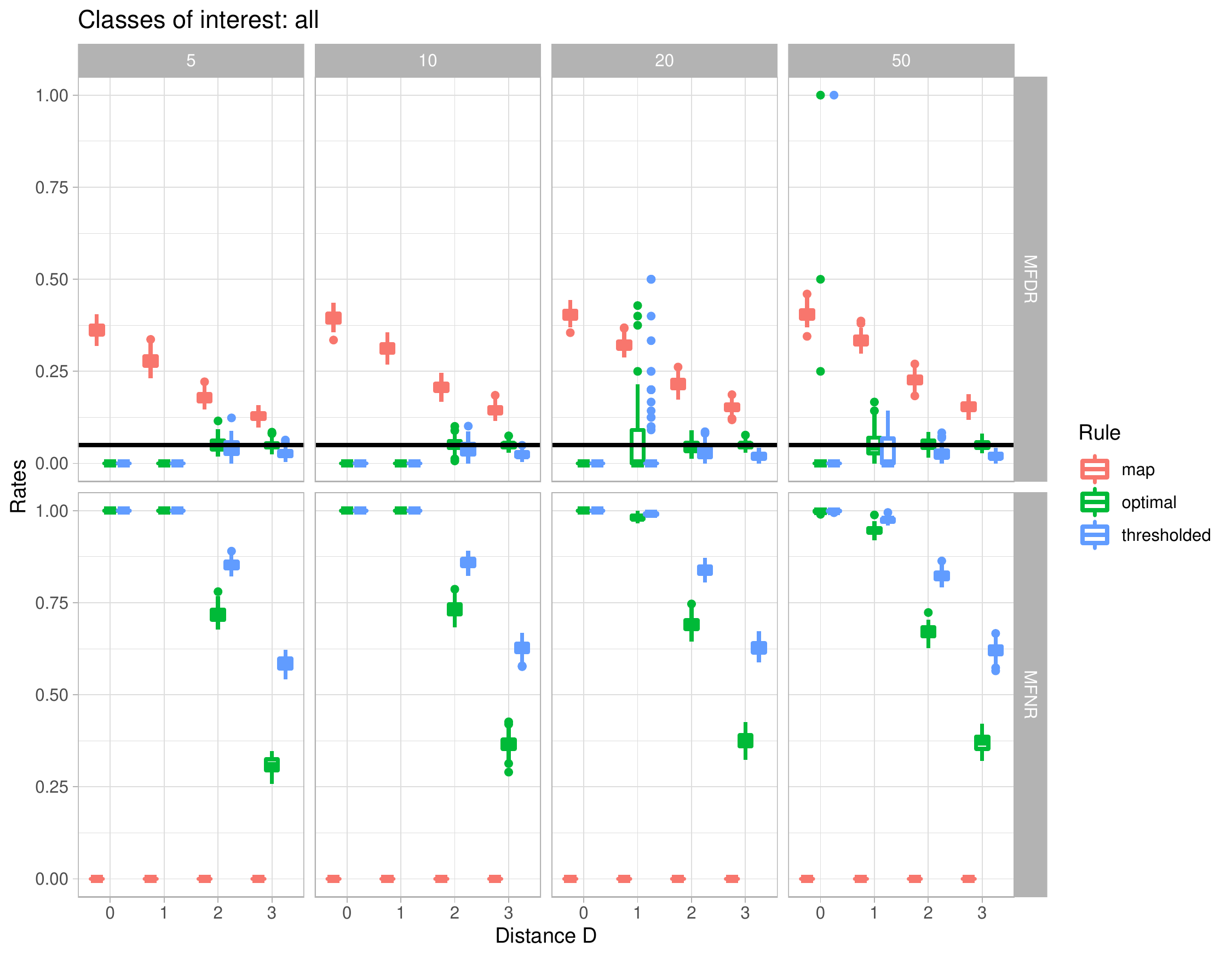}
\caption{ Performances of the MAP, optimal and $1-\alpha$ thresholded classification rules on mixture of three bivariate Student distributions. All rules are based on the true posterior probabilities. Columns correspond to different values of the degrees of freedom $\nu$. All classes are considered of interest. \label{Fig:simulations_Student_known_all}}
\end{figure}

\begin{figure}[ht!]
\includegraphics[scale=0.8]{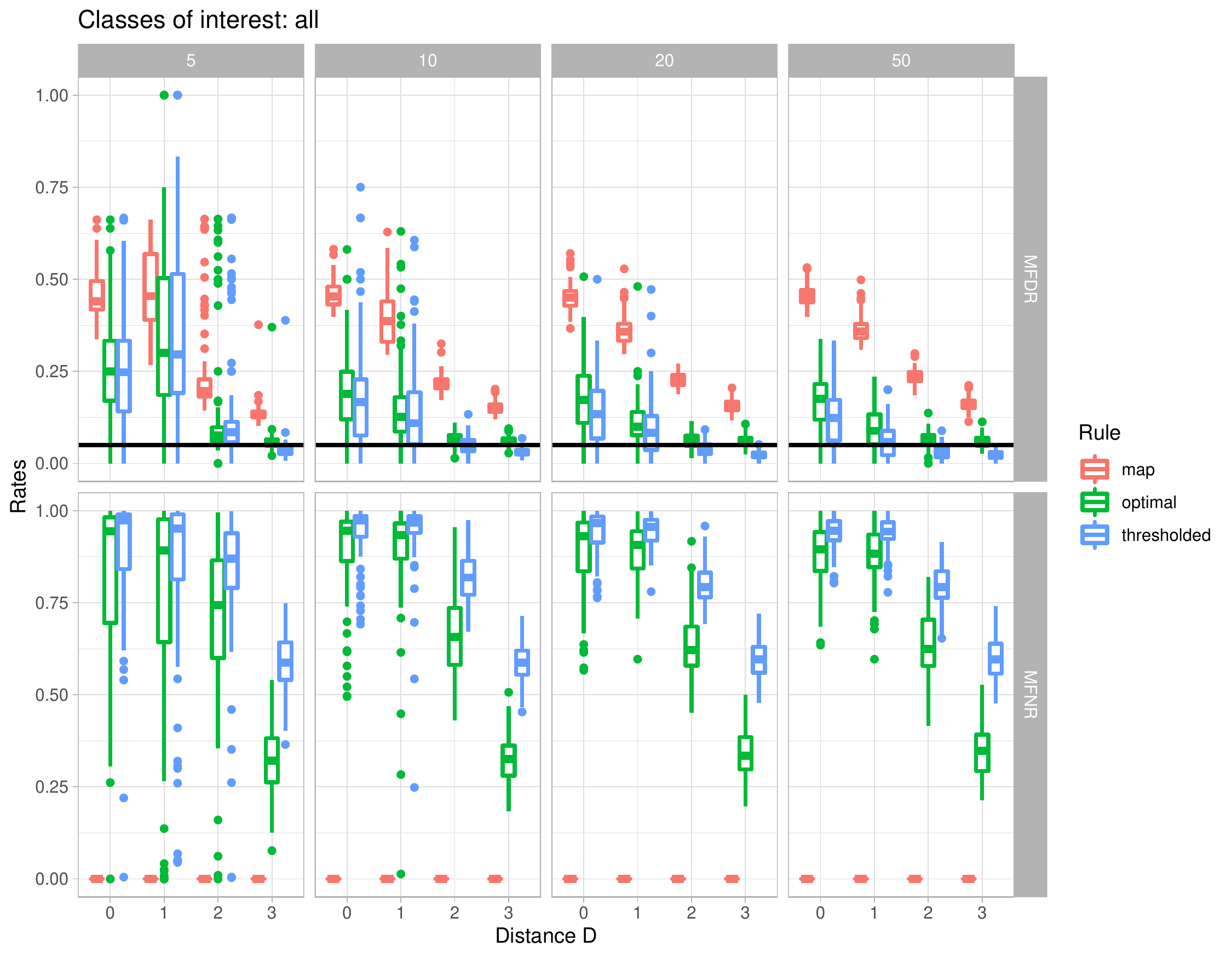}
\caption{ Same Figure as Figure \ref{Fig:simulations_Student_known_all}, except that posterior probabilities are inferred from the data. \label{Fig:simulations_Student_est_all}}
\end{figure}

\blue{Figure \ref{Fig:simulations_Student_known_all} shows the MFDR and MFNR of the considered rules using exact posterior probabilities in the case when all the classes are of interest, while Figure \ref{Fig:simulations_Student_est_all} shows error rates when the posterior probabilities are estimated using the EM algorithm for Gaussian mixtures implemented in \texttt{mclust}. Note that these results should be compared to the results shown in the central column of Figures 1 and 3 ($\sigma = 1$) in the main text.}

%\blue{(Vittorio: ajouter r\'esultats pour $K=2$? Figure avec example simulation? Boxplots avec seuils? Inverser Appendix E and Appendix F?)}

\end{document}